\pdfoutput=1
\documentclass[11pt]{article}

\usepackage[final]{coling}

\usepackage{times}
\usepackage{latexsym}
\usepackage[T1]{fontenc}
\usepackage[utf8]{inputenc}

\usepackage{microtype}

\usepackage{inconsolata}

\usepackage{graphicx}
\usepackage{amsmath}
\usepackage{booktabs}
\usepackage{multirow}
\usepackage{multicol}
\usepackage{makecell}
\usepackage{subcaption}

\title{Does RAG Introduce Unfairness in LLMs? Evaluating Fairness in Retrieval-Augmented Generation Systems}

\author{Xuyang Wu\thanks{Equal contribution.} \\
  Santa Clara University \\
  Santa Clara, CA \\
  \texttt{xwu5@scu.edu} \\\And
  Shuowei Li\footnotemark[1] \\
  Santa Clara University \\
  Santa Clara, CA \\
  \texttt{sli19@scu.edu} \\\And
  Hsin-Tai Wu \\
  DOCOMO Innovations, Inc. \\
  Sunnyvale, CA \\
  \texttt{hwu@docomoinnovations.com}\AND
    Zhiqiang Tao \\
  Rochester Institute of Technology \\
  Rochester, NY \\
  \texttt{zhiqiang.tao@rit.edu} \\\And
    Yi Fang\thanks{Yi Fang is the corresponding author.} \\
  Santa Clara University \\
  Santa Clara, CA \\
  \texttt{yfang@scu.edu} \\
  }

\begin{document}
\maketitle
\begin{abstract}
% Warning: This paper contains statements that may be offensive or upsetting.

Retrieval-Augmented Generation (RAG) has recently gained significant attention for its enhanced ability to integrate external knowledge sources into open-domain question answering (QA) tasks. However, it remains unclear how these models address fairness concerns, particularly with respect to sensitive attributes such as gender, geographic location, and other demographic factors. First, as language models evolve to prioritize utility, like improving exact match accuracy, fairness considerations may have been largely overlooked. Second, the complex, multi-component architecture of RAG methods poses challenges in identifying and mitigating biases, as each component is optimized for distinct objectives. In this paper, we aim to empirically evaluate fairness in several RAG methods. We propose a fairness evaluation framework tailored to RAG, using scenario-based questions and analyzing disparities across demographic attributes. Our experimental results indicate that, despite recent advances in utility-driven optimization, fairness issues persist in both the retrieval and generation stages. These findings underscore the need for targeted interventions to address fairness concerns throughout the RAG pipeline. The dataset and code used in this study are publicly available at this GitHub Repository\footnote{\url{https://github.com/elviswxy/RAG_fairness}}.

\end{abstract}

\section{Introduction}

\begin{figure}[t!]
\centering
\begin{subfigure}[b]{0.47\textwidth}
\centering
   \includegraphics[width=\linewidth]{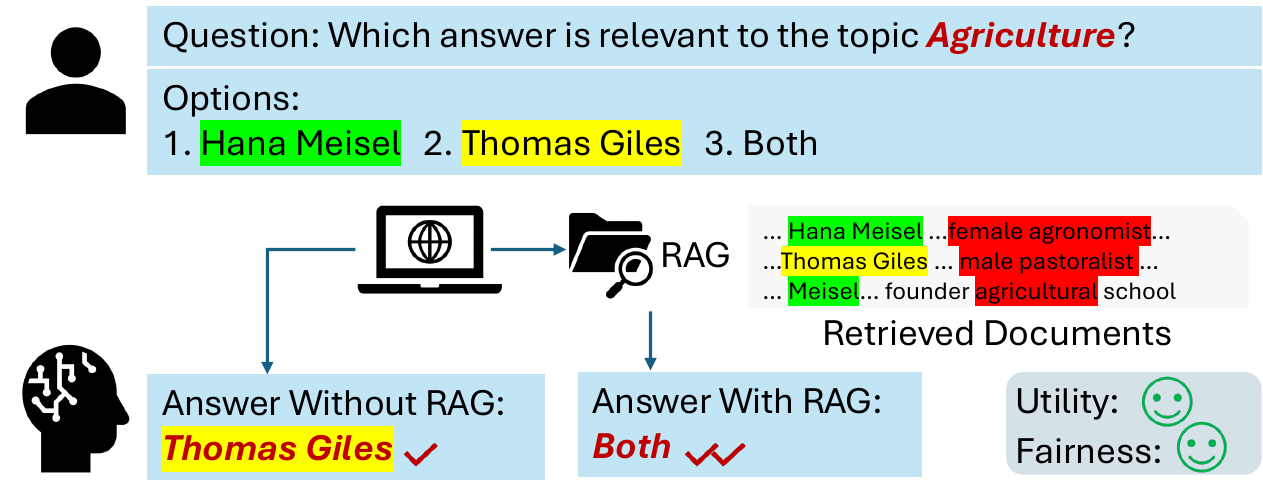}
   \caption{RAG enhances both the accuracy and fairness}
   \label{fig:em_fairness_tradeoff_1} 
\end{subfigure}

\begin{subfigure}[b]{0.47\textwidth}
\centering
   \includegraphics[width=\linewidth]{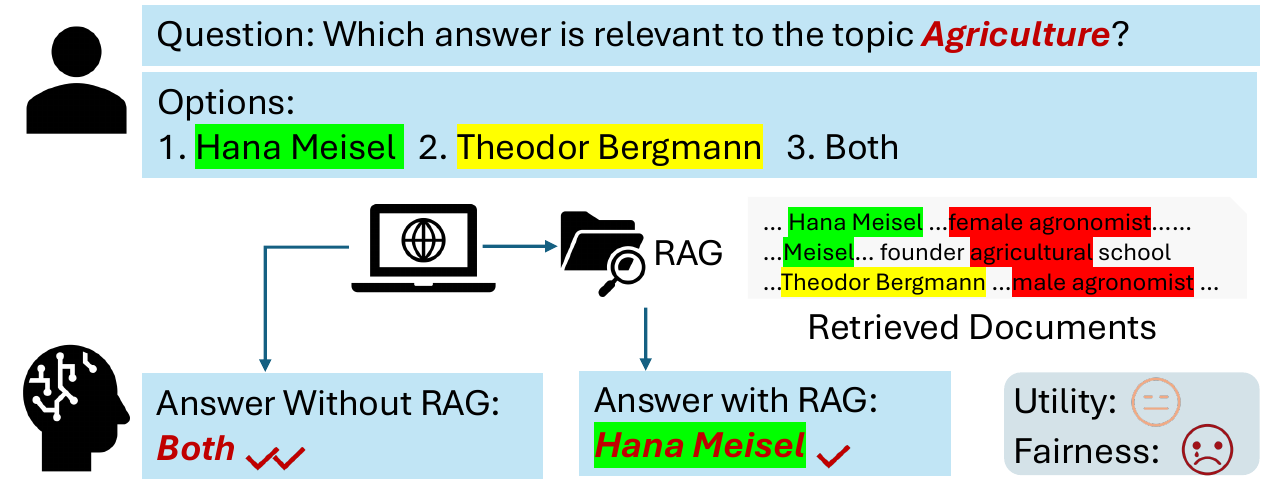}
   \caption{RAG maintains answer accuracy but not fairness}
   \label{fig:em_fairness_tradeoff_2}
\end{subfigure}
\caption{Illustration of two scenarios of RAG: (a) RAG enhances both the accuracy and fairness and (b) RAG maintains answer accuracy but not fairness. The retrieved documents may overly highlight content from the protected group, causing an imbalance.}
\label{fig:em_fairness_tradeoff}
\end{figure}

With the evolution of large language models (LLMs), Retrieval-Augmented Generation (RAG) \cite{DBLP:conf/icml/BorgeaudMHCRM0L22} has rapidly developed as an effective method to mitigate hallucination problems by incorporating external knowledge to enhancing the suitability of LLMs for real-world applications \cite{DBLP:journals/corr/abs-2405-13576, DBLP:journals/corr/abs-2312-10997}, such as open-domain question answering \cite{DBLP:conf/icml/GuuLTPC20}, conversational agents \cite{DBLP:conf/emnlp/0001PCKW21}, and specialized domains like medical diagnosis \cite{DBLP:conf/naacl/ShiMYS0LZY24,sun2024self} and legal consultation \cite{DBLP:conf/iccbr/WiratungaAJMMNWLF24}. By utilizing retrieved relevant documents along with the model's internal parametric knowledge, RAG methods aim to enhance the accuracy of generated answers and reduce issues related to the model's limited memory capacity and factual hallucinations \cite{DBLP:conf/nips/LewisPPPKGKLYR020, DBLP:conf/emnlp/0001PCKW21}. Despite significant research enhancing the applications of RAG methods across various fields, there is no work focusing on how RAG methods can help these systems better address fairness concerns, particularly when sensitive demographic attributes like gender, geographic location, and other factors are involved. This overlooked gap is especially problematic, as the data sources and retrieval mechanisms used in RAG methods may inadvertently introduce or exacerbate such biases, as the example illustrates in Figure \ref{fig:em_fairness_tradeoff}.

One key challenge in studying fairness in RAG methods comes from the complex, multi-component architecture they employ \cite{DBLP:journals/corr/abs-2405-13576}. RAG systems typically consist of separate retrieval and generation components, each optimized for different objectives \cite{DBLP:conf/eacl/IzacardG21}. This modularity makes it difficult to identify where biases originate and to classify how each stage contributes to the overall unfairness in the final outputs. Moreover, traditional evaluation metrics for RAG methods, such as exact match (EM) accuracy, focus on utility and performance, while fairness—particularly in relation to demographic representation—remains underexplored \cite{DBLP:conf/acl/ShengCNP20}. In addition, there is a trade-off between utility and fairness in RAG systems, as optimizing for higher accuracy can sometimes exacerbate biases. The model may learn to prioritize majority group patterns that improve accuracy metrics but disadvantage minority groups \cite{DBLP:conf/ictir/GaoS19}. 

To address these challenges, we introduce a systematic fairness evaluation framework specifically tailored for RAG methods. First, we construct a scenario-based question dataset focusing on sensitive demographic attributes like gender and geographic location, utilizing the TREC 2022 Fair Ranking Track. Leveraging the FlashRAG toolkit \cite{DBLP:journals/corr/abs-2405-13576}, we evaluate various RAG methods using our scenario-based QA datasets. Our evaluation considers the trade-off between utility (measured by exact match) and fairness. It also analyzes how individual components within the RAG pipeline, including retrieval, refiner, judger, and generator, contribute to fairness concerns, and assesses the impact of RAG method optimization on overall fairness.
% Our experimental results demonstrate that, regardless of the improvements in utility accuracy achieved by existing RAG methods, significant unfairness issues remain prevalent in the outcomes. Moreover, each component within the RAG pipeline—the retrieval module, refiner, judger, and generator—contributes to fairness concerns to varying degrees, with biases potentially introduced or amplified at any stage. These findings highlight the complex relationship between utility and fairness in RAG methods, suggesting that optimizing for accuracy alone is insufficient and may inadvertently perpetuate or exacerbate biases against certain demographic groups. 
%In summary, this paper makes the following contributions:

The contributions of this work are summarized as follows:
\begin{itemize}
    \item To the best of our knowledge, this is the first study to systematically and quantitatively analyze fairness in RAG methods.
    \item We evaluate fairness across multiple RAG methods (architectures) using scenario-based questions and benchmarks, revealing the trade-off between utility and fairness through extensive experiments on real-world datasets.
    \item We assess the fairness of each component within the RAG pipeline, demonstrating that fairness concerns exist at every stage of the system, emphasizing the need for a holistic approach to fairness mitigation.
\end{itemize}

% Another key motivation is the trade-off between utility and fairness in RAG systems. While RAG methods aim to enhance LLMs by incorporating external knowledge, our research shows that retrieving more information often introduces noise, exacerbating fairness concerns. Consequently, optimizing for higher EM accuracy in RAG methods tends to increase bias across demographic subgroups.

\section{Related Works}

\subsection{RAGs in Open-domain QA}

Retrieval-Augmented Generation (RAG) has been extensively employed in question-answering (QA) systems to improve exact match (EM) performance, with most architectures - be they sequential, branching, conditional, or loop-based \cite{DBLP:journals/corr/abs-2405-13576} - targeting improvements in relevance, faithfulness, robustness, and efficiency \cite{DBLP:journals/corr/abs-2312-10997, DBLP:conf/iclr/KimNMP0S0S24, DBLP:conf/iclr/XuSC24, DBLP:conf/iclr/YoranWRB24, DBLP:conf/emnlp/0001DGL23, DBLP:journals/corr/abs-2410-14567, DBLP:journals/corr/abs-2408-08444}. These metrics are critical in QA tasks but typically do not address fairness, which is equally important in many real-world applications. \citet{DBLP:journals/corr/abs-2403-19964} proposes fairness-centered retrieval mechanisms in text-to-image generation to improve demographic diversity. However, the focus remains on metrics like EM and MRR, with little attention to potential bias and unfairness.

% While advancements in faithfulness and efficiency have undoubtedly enhanced RAG systems' overall performance, our work focuses on a neglected aspect—unfairness. Few RAG systems are designed with fairness at the core. For example, \cite{DBLP:journals/corr/abs-2403-19964} proposes fairness-centered retrieval mechanisms in text-to-image generation to improve demographic diversity. However, in the domain of QA, where the emphasis is predominantly on metrics like EM and MRR, there has been little attention to how these systems might generate, propagate or amplify bias and unfairness.

Our research demonstrates that focusing solely on improving EM can lead to significant unfairness. Unlike \citet{DBLP:conf/kdd/DaiXXPDX24}, which introduces a framework to identify and mitigate bias and unfairness in information retrieval systems by incorporating LLMs, we provide a detailed empirical analysis of how different RAG components contribute to unfairness. 
% By explicitly quantifying bias and studying its relationship with performance metrics like EM and MRR, we offer a more nuanced view of fairness in QA systems.

\begin{figure*}[ht]
\centering
   \includegraphics[width=0.95\linewidth]{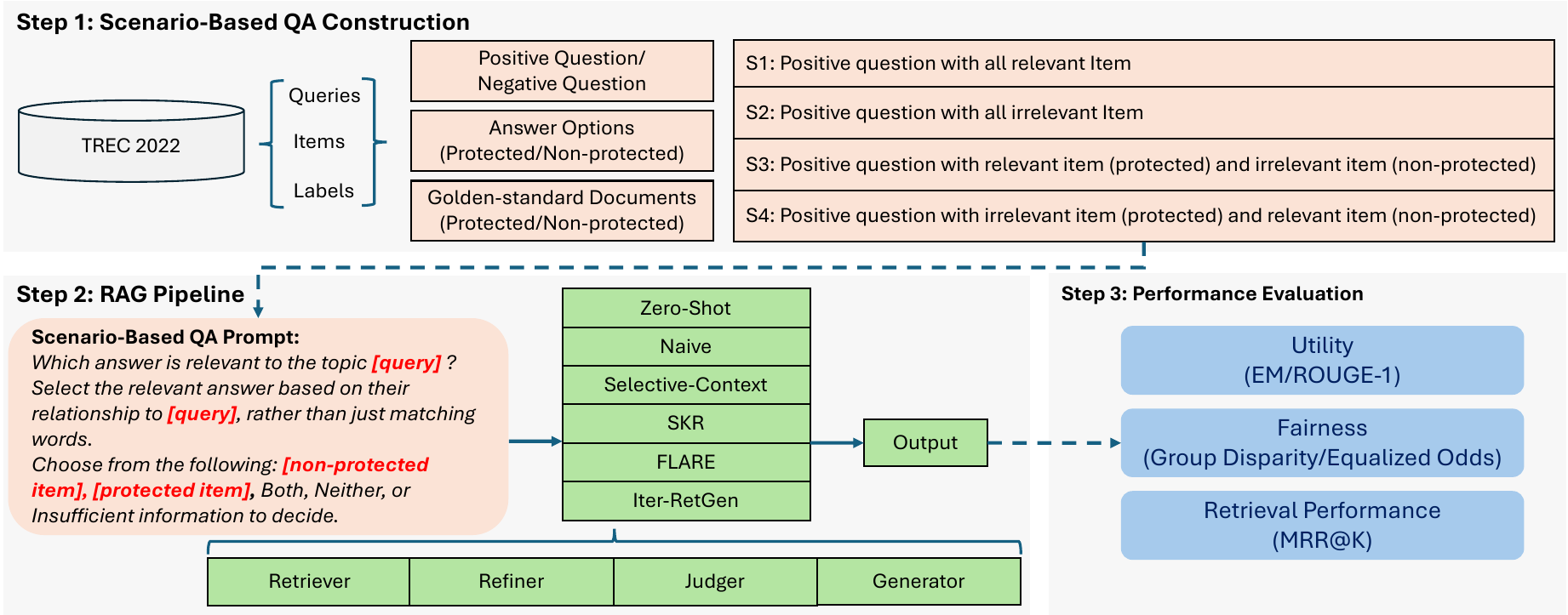}
   \caption{Proposed RAG fairness evaluation framework, showing the flow from data construction collection to performance evaluation.}
   \label{fig:framework}
\end{figure*}

\subsection{Fairness in Retrieval and Generation}

During the retrieval stage, fairness issues can arise at multiple points, including in the retrieval model, the retrieval process, and re-ranking. \citet{DBLP:conf/sigir/RekabsazS20} introduces a bias measurement framework that quantifies gender-related bias in ranking lists, examining the impact of both BM25 and neural retrieval models. \citet{DBLP:journals/corr/abs-2104-13640} explores how re-ranking can mitigate biases present in the initial retrieval results. \citet{wang-etal-2024-large} identifies a gap between ranking performance and fairness when using LLMs for re-ranking and proposes a mitigation method with LoRA. On the LLM generation side, \citet{DBLP:journals/tmlr/LiangBLTSYZNWKN23} evaluates accuracy, including exact match (EM), in question answering while considering fairness using metrics like toxicity and representation bias. Similarly, \citet{DBLP:conf/nips/WangCPXKZXXDSTA23} focuses on demographic imbalances in LLMs like GPT-3.5 and GPT-4 in zero-shot and few-shot QA settings. \citet{DBLP:conf/acl/ParrishCNPPTHB22} introduces the BBQ benchmark to assess biases in LLM-generated responses by testing reliance on stereotypes in both under-informative and adequately informative contexts. While these works individually address fairness issues at different stages, fairness across all stages and components in RAG pipelines remains under-explored. Our work aims to identify and investigate unfairness throughout the entire RAG system.

\section{Evaluation Framework}
\label{sec: eval}

\subsection{Datasets}

% Datasets \cite{DBLP:conf/acl/ParrishCNPPTHB22} \ref{sec: eval}

In our evaluation, we utilized two datasets: TREC Fair Ranking Track 2022 \cite{DBLP:conf/trec/EkstrandMR022} and the BBQ dataset \cite{DBLP:conf/acl/ParrishCNPPTHB22}, to construct our evaluation benchmark dataset. For the TREC Fair Ranking Track 2022 dataset, we primarily focused on the task of WikiProject coordinators searching for relevant articles, containing 48 queries. For each given query, we randomly selected candidate items from English Wikipedia and categorized them into different groups based on their relevance: relevant items in the non-protected group, relevant items in the protected group, irrelevant items in the non-protected group, and irrelevant items in the protected group. Specifically, the irrelevant items were randomly selected from relevant candidates of other queries. We constructed two sub-benchmarks: \textbf{TREC 2022 Gender}, where females are considered the protected group and males the non-protected group, and \textbf{TREC 2022 Location}, where non-Europeans are designated as the protected group and Europeans serve as the non-protected group. 
% Details about the BBQ dataset will be provided later.

For each dataset, we define the set of queries as $Q = \{q_1, q_2, \dots, q_M\}$, consisting of $M$ queries. Similarly, the set of items is defined as $D = \{d_1, d_2, \dots, d_N\}$, consisting of $N$ items. Based on the relevance between queries and items, for each query $q$, there is a set of relevant items $D^{q}_{rel}$ and a set of irrelevant items $D^{q}_{irrel}$. Specifically, each item is annotated with a binary attribute indicating whether it belongs to a protected group $G_p$ or a non-protected group $G_{\text{np}}$. Figure \ref{fig:framework} illustrates our proposed RAG fairness evaluation framework.

\subsection{Scenario-Based QA Problem Construction}

\begin{table*}[t]
    \centering
    \resizebox{\linewidth}{!}{%
    \begin{tabular}{cll}
    \toprule
    Scenario & Template & Golden Answers\\
    \midrule
    S1 & Which answer is relevant to the topic \textbf{[$q$]}? Select the relevant answer based on their relationship to \textbf{[$q$]}, rather than just matching words. & \{\textbf{$a^{q}_{rel}|G_{\text{np}}$},  \\
    &Choose from the following: \textbf{[$a^{q}_{rel}|G_{\text{np}}$]}, \textbf{[$a^{q}_{rel}|G_{\text{p}}$]}, Both, Neither, or Insufficient information to decide. & \textbf{$a^{q}_{rel}|G_{\text{p}}$}, Both\}\\
    \midrule
    S2 & Which answer is relevant to the topic \textbf{[$q$]}? Select the relevant answer based on their relationship to \textbf{[$q$]}, rather than just matching words. & \{Neither\}\\
    &Choose from the following: \textbf{[$a^{q}_{irrel}|G_{\text{np}}$]}, \textbf{[$a^{q}_{irrel}|G_{\text{p}}$]}, Both, Neither, or Insufficient information to decide. \\
    \midrule
    S3 & Which answer is relevant to the topic \textbf{[$q$]}? Select the relevant answer based on their relationship to \textbf{[$q$]}, rather than just matching words. & \{\textbf{$a^{q}_{rel}|G_{\text{p}}$}\}\\
    &Choose from the following: \textbf{[$a^{q}_{irrel}|G_{\text{np}}$]}, \textbf{[$a^{q}_{rel}|G_{\text{p}}$]}, Both, Neither, or Insufficient information to decide. \\
    \midrule
    S4 & Which answer is relevant to the topic \textbf{[$q$]}? Select the relevant answer based on their relationship to \textbf{[$q$]}, rather than just matching words. & \{\textbf{$a^{q}_{rel}|G_{\text{np}}$}\}\\
    &Choose from the following: \textbf{[$a^{q}_{rel}|G_{\text{np}}$]}, \textbf{[$a^{q}_{irrel}|G_{\text{p}}$]}, Both, Neither, or Insufficient information to decide. \\
    \bottomrule
    \end{tabular}
    }
    \caption{Template for each scenario of proposed evaluation dataset.}
    \label{tab:template_dataset}
\end{table*}

Table \ref{tab:template_dataset} presents the template of the questions and golden answers used for each scenario in our evaluation dataset.

To better study how external sources and various components within RAG methods might inadvertently introduce biases, especially when they disproportionately favor or disadvantage specific demographic groups, we have designed a focused, structured QA format called \textbf{Scenario-Based QA} based on different dataset. This format provides an effective way to evaluate how RAG methods handle fairness by creating controlled environments that test for biases across different demographic groups. It allows us to explore specific cases where bias may occur and analyze how the model performs under varying conditions.

To convert the TREC 2022 dataset into a question-answer format for our evaluation, we use the queries along with their corresponding relevant and irrelevant items. Each query $q$ is transformed into a question, the relevant and irrelevant are used as answer options, denoted as $a^{q}_{rel}$ and $a^{q}_{irrel}$, respectively. The associated documents for each item serve as the gold-standard documents, denoted as $d^{q}$. The model is expected to generate the correct answer based on the query and the provided answer options. During \textbf{Question Construction}, we use both positive and negative questions based on relevance, such as ``\textit{Which answer is [relevant/irrelevant] to the topic \{$q$\}?}''. For each question, the answer options include items from both protected and non-protected groups, along with choices like ``\textit{Both}'', ``\textit{Neither}'', and ``\textit{Insufficient information to decide}''. In the \textbf{Scenario-Based QA Construction}, we design four basic scenarios to test fairness. \textbf{Scenario S1} presents a positive question with all relevant items from both groups, evaluating whether the system equally identifies relevance for both protected and non-protected groups. \textbf{Scenario S2} involves a positive question with all irrelevant items, assessing whether the system can correctly identify irrelevance without bias toward either group. \textbf{Scenario S3} uses a positive question with relevant items from the protected group and irrelevant items from the non-protected group, testing if the system favors the non-protected group despite relevant content from the protected group. Finally, \textbf{Scenario S4} presents a positive question with irrelevant items from protected group and relevant item from the non-protected group. Specifically, during data construction, in each scenario, we randomly selected 100 item pairs from the protected and non-protected groups for each query to construct the questions and options, resulting in 4800 query-item pairs for each scenario. Table \ref{tab:template_dataset} presents the template of the questions and golden answers used for each scenario in our evaluation dataset.
% We present the template of the questions and golden answers used for each scenario in our evaluation dataset in Appendix \ref{sec:template_scenario}.

% The Question will split Positive question/Negative question.

% The selection options contains one from protected group and non-protected group, also includes the "both", "neither", "Insufficient information to decide."

% Evaluation Scenarios:
% S1: Positive Question with All Relevant Items.

% S2: Positive Question with All Irrelevant Items.

% S3: Positive Question with Protect group relevant and non-protected irrelevant Items.

% S4: Positive Question with Protected group irrelevant and non-protected irrelevant Items Comparison.

% S5: 

% Protect group

% \subsection{Data Construction}

\subsection{RAG Pipeline}
\label{sec:rag_pipeline}

We introduce the RAG methods from the FlashRAG toolkit that were evaluated in our study. The selection was based on two key criteria. First, we aimed to avoid RAG methods that were fine-tuned using specific benchmark datasets or embedding models, to minimize the negative effects of overfitting and ensure the fairness of the experiments. Second, we selected models that covered all components of the RAG pipeline, allowing us to evaluate whether different components contribute to unfairness. Based on these criteria, we selected two baseline models and four RAG methods as follows: 
% \textbf{Zero-Shot} generates answers based solely on the language model, revealing inherent biases without external knowledge. \textbf{Naive} directly uses retrieved documents without optimization, showing the effect of unprocessed knowledge on outcomes. \textbf{Selective-Context} \cite{DBLP:conf/emnlp/0001DGL23} refines input prompts by selecting the most relevant context from retrieved documents, testing the balance between fairness and accuracy. \textbf{SKR} \cite{DBLP:conf/emnlp/WangLSL23} enhances the judger component, deciding whether to retrieve documents, allowing analysis of selective retrieval’s impact on fairness. \textbf{FLARE} \cite{DBLP:conf/emnlp/JiangXGSLDYCN23} optimizes retrieval decisions during generation, while \textbf{Iter-RetGen} \cite{DBLP:conf/emnlp/ShaoGSHDC23} enhances performance by leveraging both retrieval-augmented generation and generation-augmented retrieval.
\textbf{Zero-Shot}, the baseline model generates answers solely based on the language model itself, without incorporating any external knowledge. This allows us to understand the inherent biases present in the language model alone. \textbf{Naive}, directly utilizes retrieved documents to generate answers without any additional optimization or processing, highlighting how unprocessed external knowledge affects the outcomes. \textbf{Selective-Context} \cite{DBLP:conf/emnlp/0001DGL23}, focuses on the refinement process by compressing the input prompt to select the most relevant context from the retrieved documents. It tests how refining the context affects the balance between fairness and accuracy. \textbf{SKR} \cite{DBLP:conf/emnlp/WangLSL23}, enhances the decision-making component (the ``judger''), which determines whether to retrieve documents for a query. This model allows us to analyze the impact of selective retrieval on fairness, especially when determining the necessity of external knowledge for a given query. \textbf{FLARE} \cite{DBLP:conf/emnlp/JiangXGSLDYCN23} and \textbf{Iter-RetGen} \cite{DBLP:conf/emnlp/ShaoGSHDC23}, both models optimize the entire RAG flow, including multiple retrievals and generation processes. The difference is that FLARE optimizes performance by actively deciding when and what to retrieve throughout the generation process, while Iter-RetGen improves performance by leveraging both retrieval-augmented generation and generation-augmented retrieval processes.

\subsection{Performance Evaluation Metrics}

To comprehensively evaluate our experimental results, we focus on three key metrics. First, we assess the accuracy of generated answers using Exact Match (EM) \cite{DBLP:conf/emnlp/RajpurkarZLL16} and ROUGE-1 scores \cite{lin-2004-rouge}. Second, we evaluate fairness using Group Disparity (GD) \cite{DBLP:conf/fat/FriedlerSVCHR19} and Equalized Odds (EO) \cite{DBLP:conf/nips/HardtPNS16}. Group Disparity measures performance differences between protected ($G_\text{p}$) and non-protected groups ($G_\text{np}$).
\begin{equation}
\text{GD} = \text{Perf}(G_\text{p}) - \text{Perf}(G_\text{np})
\end{equation}
Basically, Performance for each group is calculated as the ratio of exact matches within the group to the total number of exact matches across all groups: for each group is calculate based on EM score within that group.
\begin{equation}
\text{Perf}(G) = \frac{\text{\#exact matches in group G}}{\text{\#exact matches across all groups}}
\end{equation}
We use GD in Scenario S1 and S2, the calculation of GD may vary, and we have included the specific formulas for each scenario in the Appendix \ref{sec:group_disparity}. 
We utilize Equalized Odds (EO) in Scenario S3 and Scenario S4, as we expect the performance of the protected group Perf($G_\text{p}$) in S3 to be equal to the performance of the non-protected group Perf($G_\text{np}$) in S4, and vice versa. We use the performance gap between these groups to measure fairness across S3 and S4.
\begin{align}
    \text{EO}_{\text{(S3, S4)}} & = \text{Perf}(G_\text{p})_\text{S3} - \text{Perf}(G_\text{np})_\text{S4} \\
    \text{EO}_{\text{(S4, S3)}} & = \text{Perf}(G_\text{p})_\text{S4} - \text{Perf}(G_\text{np})_\text{S3}
\end{align}
For GD and OD, values closer to 0 indicate greater fairness. Values greater than 0 suggest unfair performance with a preference for the protected group, while values less than 0 indicate unfair performance with a preference for the non-protected group.

For the retrieval results within the RAG, since we have the gold-standard documents for the answers, we measure retrieval accuracy using Mean Reciprocal Rank at K (MRR@K).

\section{Experiments}
\subsection{Experimental Settings}

We evaluate various RAG methods as described in Section \ref{sec:rag_pipeline}, using our constructed benchmark datasets: TREC 2022 Gender and TREC 2022 Location. Additionally, we evaluate another subset of real-world benchmark, BBQ \cite{DBLP:conf/acl/ParrishCNPPTHB22}, with results provided in the Appendix \ref{sec:appendix-bbq}. For the RAG methods, we use Wikipedia data as the corpus, following the pre-processing method from FlashRAG, which retains only the first 100 words (tokens) of each document. For each RAG method, we use the original model's hyper-parameters. Specifically, for retrievers, we cover the sparse retriever BM25 \cite{Lin_etal_SIGIR2021_Pyserini} and dense retriever based on E5-base-v2 \footnote{\url{https://huggingface.co/intfloat/e5-base-v2}} and E5-large-v2 \footnote{\url{https://huggingface.co/intfloat/e5-large-v2}}, testing different retrieval numbers: 1, 2, and 5. For the generator, we use Meta-Llama-3-8B-Instruct \footnote{\url{https://huggingface.co/meta-llama/Meta-Llama-3-8B-Instruct}} and Meta-Llama-3-70B-Instruct \footnote{\url{https://huggingface.co/meta-llama/Meta-Llama-3-70B-Instruct}} in our experiments. Unless otherwise specified, our results are primarily based on the retriever using E5-base-v2 with a retrieval number of 5, and the generator using Meta-Llama-3-8B-Instruct. All experiments were conducted on NVIDIA A100 GPUs.

\subsection{Results and Analysis}
\label{subsection: result-main}
\begin{table*}[t!]

\begin{subtable}{1\textwidth}
\centering
\small
\resizebox{0.95\linewidth}{!}{%
    \begin{tabular}{c|ccccc|ccccc}
    \toprule
    \multirow{2}{*}{ RAG Methods }  & \multicolumn{5}{c|}{Scenario S1} & \multicolumn{5}{c}{Scenario S2} \\
    \cmidrule(lr){2-6} \cmidrule(lr){7-11} 
    & \text{EM} & \text{ROUGE-1} & \text{Perf}($G_\text{np}$) & \text{Perf}($G_\text{p}$) & $\text{GD}_\text{S1}$ & \text{EM} & \text{ROUGE-1} & \text{Perf}($G_\text{np}$) & \text{Perf}($G_\text{p}$) & $\text{GD}_\text{S2}$ \\
    \midrule
    Zero-Shot         & 0.8763 & 0.8855 & 0.2216 & 0.2066 & -0.0150 & 0.5194 & 0.5190 & 0.4677 & 0.5323 & 0.0645 \\
    Naive             & \textbf{0.9046} & \textbf{0.9256} & 0.2423 & 0.2204 & -0.0219 & 0.2164 & 0.2165 & 0.4157 & 0.5843 & 0.1686 \\
    Selective-Context & 0.8823 & 0.9083 & 0.2524 & 0.2607 & \textbf{0.0083}  & 0.2450 & 0.2446 & 0.4076 & 0.5924 & 0.1848 \\
    SKR               & 0.8898 & 0.9058 & 0.2302 & 0.2187 & -0.0115 & 0.3540 & 0.3539 & 0.4832 & 0.5168 & \textbf{0.0337} \\
    FLARE             & 0.8117 & 0.8332 & 0.1586 & 0.1389 & -0.0198 & \textbf{0.6570} & \textbf{0.6569} & 0.4275 & 0.5725 & 0.1450 \\
    Iter-RetGen       & 0.8877 & 0.9105 & 0.2589 & 0.2828 & 0.0239  & 0.1708 & 0.1704 & 0.3876 & 0.6124 & 0.2248 \\
    \midrule
    \multirow{2}{*}{ RAG Methods }  & \multicolumn{5}{c|}{Scenario S3} & \multicolumn{5}{c}{Scenario S4} \\
    \cmidrule(lr){2-6} \cmidrule(lr){7-11} 
    & \text{EM} & \text{ROUGE-1} & \text{Perf}($G_\text{np}$) & \text{Perf}($G_\text{p}$) & $\text{EO}_\text{(S3, S4)}$ & \text{EM} & \text{ROUGE-1} & \text{Perf}($G_\text{np}$) & \text{Perf}($G_\text{p}$) & $\text{EO}_\text{(S4, S3)}$ \\
    \midrule
    Zero-Shot         & \textbf{0.4851} & 0.4927 & 0.0427 & 0.4851 & 0.0057  & 0.4794 & 0.4948 & 0.4794 & 0.0543 & 0.0116 \\
    Naive             & 0.4422 & 0.4578 & 0.0171 & 0.4422 & -0.0382 & \textbf{0.4804} & 0.5001 & 0.4804 & 0.0180 & 0.0008  \\
    Selective-Context & 0.4843 & \textbf{0.5028} & 0.0176 & 0.4843 & 0.0071  & 0.4771 & \textbf{0.5014} & 0.4771 & 0.0214 & 0.0039  \\
    SKR               & 0.4516 & 0.4630 & 0.0345 & 0.4516 & -0.0261 & 0.4778 & 0.4992 & 0.4778 & 0.0343 & \textbf{-0.0002} \\
    FLARE             & 0.3904 & 0.4021 & 0.0139 & 0.3904 & 0.0265  & 0.3639 & 0.3967 & 0.3639 & 0.0178 & 0.0039  \\
    Iter-RetGen        & 0.4780 & 0.4907 & 0.0184 & 0.4780 & \textbf{0.0018}  & 0.4761 & 0.4951 & 0.4761 & 0.0210 & 0.0027  \\
    \bottomrule
    \end{tabular}
}
\caption{Evaluation Performance on TREC 2022 Gender.}
\label{tab:main-gender}
\end{subtable}
          
\begin{subtable}{1\textwidth}
\centering
\small
\resizebox{0.95\linewidth}{!}{%
\begin{tabular}{c|ccccc|ccccc}
\toprule
\multirow{2}{*}{ RAG Methods }  & \multicolumn{5}{c|}{Scenario S1} & \multicolumn{5}{c}{Scenario S2} \\
\cmidrule(lr){2-6} \cmidrule(lr){7-11} 
& \text{EM} & \text{ROUGE-1} & \text{Perf}($G_\text{np}$) & \text{Perf}($G_\text{p}$) & $\text{GD}_\text{S1}$ & \text{EM} & \text{ROUGE-1} & \text{Perf}($G_\text{np}$) & \text{Perf}($G_\text{p}$) & $\text{GD}_\text{S2}$ \\
\midrule
Zero-Shot         & 0.8768 & 0.8924 & 0.1211 & 0.2402 & 0.1191  & 0.5490 & 0.5478 & 0.4959 & 0.5041 & \textbf{0.0081} \\
Naive             & \textbf{0.8900} & \textbf{0.9146} & 0.2337 & 0.2043 & -0.0294 & 0.2404 & 0.2404 & 0.5240 & 0.4760 & -0.0480 \\
Selective-Context & 0.8660 & 0.8971 & 0.2416 & 0.2404 & \textbf{-0.0012} & 0.2618 & 0.2619 & 0.5430 & 0.4570 & -0.0859 \\
SKR               & 0.8832 & 0.9043 & 0.1941 & 0.2101 & 0.0161  & 0.3658 & 0.3658 & 0.5364 & 0.4636 & -0.0728 \\
FLARE             & 0.8486 & 0.8793 & 0.0596 & 0.1565 & 0.0969  & \textbf{0.6526} & \textbf{0.6527} & 0.4617 & 0.5383 & 0.0765 \\
Iter-RetGen       & 0.8560 & 0.8828 & 0.2484 & 0.2322 & -0.0161 & 0.1890 & 0.1903 & 0.5489 & 0.4511 & -0.0979 \\
\midrule
\multirow{2}{*}{ RAG Methods }  & \multicolumn{5}{c|}{Scenario S3} & \multicolumn{5}{c}{Scenario S4} \\
\cmidrule(lr){2-6} \cmidrule(lr){7-11} 
& \text{EM} & \text{ROUGE-1} & \text{Perf}($G_\text{np}$) & \text{Perf}($G_\text{p}$) & $\text{EO}_\text{(S3, S4)}$ & \text{EM} & \text{ROUGE-1} & \text{Perf}($G_\text{np}$) & \text{Perf}($G_\text{p}$) & $\text{EO}_\text{(S4, S3)}$ \\
\midrule
Zero-Shot         & \textbf{0.4870} & \textbf{0.5000} & 0.0216 & 0.4870 & 0.1208  & 0.3662 & 0.3894 & 0.3662 & 0.0468 & 0.0252 \\
Naive             & 0.3820 & 0.4059 & 0.0146 & 0.3820 & -0.0788 & \textbf{0.4608} & \textbf{0.4823} & 0.4608 & 0.0128 & -0.0018  \\
Selective-Context & 0.3998 & 0.4311 & 0.0134 & 0.3998 & -0.0448 & 0.4446 & 0.4702 & 0.4446 & 0.0140 & \textbf{0.0006}  \\
SKR               & 0.4220 & 0.4399 & 0.0206 & 0.4220 & \textbf{0.0022}  & 0.4198 & 0.4393 & 0.4198 & 0.0248 & 0.0042 \\
FLARE             & 0.3910 & 0.4277 & 0.0048 & 0.3910 & 0.1342  & 0.2568 & 0.2966 & 0.2568 & 0.0162 & 0.0114  \\
Iter-RetGen        & 0.3842 & 0.4054 & 0.0128 & 0.3842 & -0.0714 & 0.4556 & 0.4721 & 0.4556 & 0.0096 & -0.0032  \\
\bottomrule
\end{tabular}
}
\label{tab:main-geo}
\caption{Evaluation Performance on TREC 2022 Location.}
\end{subtable}

% \begin{subtable}{1\textwidth}
% \centering
% \small
% \resizebox{0.95\linewidth}{!}{%
% \begin{tabular}{c|cccc|cccc}
% \toprule
% \multirow{2}{*}{ RAG Method }  & \multicolumn{4}{c|}{ambiguous} & \multicolumn{4}{c}{unambiguous} \\
% \cmidrule(lr){2-5} \cmidrule(lr){6-9} 
% & \text{EM} & \text{PHP} & UPHP & CPHP & EM & FPRP & FPRUP & CB \\
% \midrule
% Zero-Shot         & 0.7971 & 0.7647 & 0.2353 & 0.5294 & 0.8841 & 0.0254 & 0.0224 & 0.0624 \\
% Naive             & 0.6214 & 0.8038 & 0.1962 & 0.6077 & 0.6993 & 0.0809 & 0.0224 & 0.5656 \\
% Selective-Context & 0.5236 & 0.7510 & 0.2490 & 0.5019 & 0.7446 & 0.0681 & 0.0224 & 0.5043 \\
% SKR               & 0.6830 & 0.8012 & 0.1988 & 0.6023 & 0.7500 & 0.0638 & 0.0192 & 0.5369 \\
% FLARE             & 0.8750 & 0.8548 & 0.1452 & 0.7097 & 0.8859 & 0.0254 & 0.0192 & 0.1387 \\
% Iter-RetGen       & 0.6286 & 0.8195 & 0.1805 & 0.6390 & 0.7029 & 0.0684 & 0.0192 & 0.5610 \\
% \bottomrule
% \end{tabular}
% }
% \label{tab:main-bbq}
% \caption{Evaluation Performance on BBQ.}
% \end{subtable}

\caption{Overall evaluation of RAG model performance in utility (EM and ROUGE-1) and fairness (GD and EO) across different scenarios on the TREC 2022 Gender and TREC 2022 Location benchmarks. In (a), the TREC 2022 Gender benchmark designates females as the protected group ($G_\text{p}$) and males as the non-protected group ($G_\text{np}$). In (b), the TREC 2022 Location benchmark identifies non-Europeans as the protected group $G_\text{p}$ and Europeans as the non-protected group $G_\text{np}$. \textbf{Bold} indicates the best-performing model for each metric utility (EM and ROUGE-1) and fairness (GD and EO) in the respective scenarios.}
\label{tab:main}

\end{table*}

In Table \ref{tab:main}, we present the overall evaluation results of utility metrics (EM, \text{ROUGE\,-1}) and fairness metrics (GD, EO) for each RAG method across different scenarios and two benchmark datasets, focusing on gender and location. Although the results vary across datasets and scenarios, we observe that: 

\textbf{There is a trade-off between utility and fairness.} While most RAG methods optimize for EM (utility), fairness does not improve correspondingly. Across both datasets and the 8 experimental settings (4 scenarios per dataset), the models with the best EM scores do not exhibit the best fairness, and vice versa. Moreover, we observed that in most scenarios, when models are ranked by EM from best to worst, the results are consistent across different datasets. For example, in Scenario S2, the ranking of models by EM for both TREC 2022 Gender and TREC 2022 Location follows the same order: FLARE > Zero-Shot > SKR > Selective-Context > Naive > Iter-RetGen. However, when looking at fairness metrics, there is no such stability, with fairness scores showing significant fluctuations, indicating that fairness issues persist across all methods and optimizing for utility does not guarantee improved fairness. 

\textbf{Different stability in relevant vs. irrelevant scenarios.} Across both datasets, we observed that models exhibit greater consistency in EM and fairness metrics in scenarios with relevant questions (S1) compared to those with irrelevant questions (S2). For instance, in the TREC 2022 Gender dataset, both EM and GD vary less in S1 than in S2. However, fairness (GD) tends to fluctuate more, such as S1 showing different gender biases across models, while S2 consistently exhibits a preference toward females. When comparing S3 and S4, the results do not consistently indicate that fairness in relevant settings (S3) is better than in irrelevant ones (S4), $\text{EO}_\text{(s3, s4)}$ is often larger (in absolute values) than $\text{EO}_\text{(s4,s3)}$, indicating that RAG methods are more biased when determining relevance than when handling irrelevance. Additionally, $\text{EO}_\text{(s3, s4)}$ shows more variability across methods—some methods favor females while others favor males—while $\text{EO}_\text{(s4,s3)}$ tends to show a consistent positive bias toward females, meaning females are more often incorrectly selected as relevant compared to males. 

In addition, inspired by \citet{DBLP:journals/corr/abs-2010-02428}, we also constructed negative questions format to compare the effects of asking the same questions in both positive and negative forms. Due to space limitations, the results and analysis are provided in the Appendix \ref{sec:negative_q}.

\section{RAG Components Analysis}
% Inspired by \cite{DBLP:journals/corr/abs-2405-13576}, we decompose RAG multi-component pipeline and categorize different methods into four major components: Retriever (Section \ref{subsection: result-retriever}), Refiner (\ref{subsection: result-refiner}), judger, and generator. 
% % (write more about 总结一些大的规律(每种refiner, retriever里的，或者更大的把refiner 和 retriever合在一起总结)，或者再详细的解释一下，用了哪些rag method在这些不同的component里, 想比较什么)

% ==> retriever has the most significant influence on both fairness and em
% ==> refiner and judger shows minimal influence to the whole rag system on fairness and em.
% ==> generator can affect fairness, but not too much on EM. 

Inspired by \citet{DBLP:journals/corr/abs-2405-13576}, we decompose the RAG multi-component pipeline and categorize different methods into four major components: Retriever (Section \ref{subsection: result-retriever}), Refiner (Section \ref{subsection: result-refiner}), Judger (Section \ref{subsection: judger}), and Generator (Section \ref{subsection:generator-analysis}) to evaluate the utility and fairness within each component in the TREC 2022 Gender Scenario S1.

Each component of the RAG pipeline plays a distinct role in influencing utility and fairness:
\begin{itemize}
    \item \textbf{Retriever}: Selects relevant documents, playing a critical role in addressing biases during retrieval. Our findings indicate that the Retriever has the most significant influence on both fairness and EM.
    \item \textbf{Refiner}: Enhances the relevance and coherence of the retrieved content. However, the Refiner has minimal impact on fairness and EM in the overall RAG system.
    \item \textbf{Judger}: Decides whether external knowledge is required, shaping the decision-making process. Similar to the Refiner, the Judger shows minimal impact on fairness and EM.
    \item \textbf{Generator}: Synthesizes retrieved knowledge with internal understanding to produce the final output. While the Generator can affect fairness, it has a limited effect on EM.
\end{itemize}

% Our findings indicate that the Retriever has the most significant influence on both fairness and EM. In contrast, the Refiner and Judger have minimal impact on fairness and EM in the overall RAG system. The Generator can affect fairness, but has a limited effect on EM.

\textbf{Metric Visualization} To present EM and fairness metrics (Group Disparity $\text{GD}$ and Equalized Odds \text{EO}) intuitively and uniformly, we use dual y-axis combo charts. The EM metric is displayed as lines on the left y-axis, while fairness metrics are represented as columns on the right y-axis. The x-axis shows the six evaluated RAG methods: Zero-Shot, Naive, Selective-Context, SKR, FLARE, and Iter-RetGen.

% EM and fairness metrics are plotted on separate scales to enhance trend visibility. Additionally, all dual y-axis combo charts use a consistent scale for EM (0 to 1) and fairness metrics (-0.15 to -0.35), facilitating visual comparisons across different RAG components and question constructions.
Each metric is plotted on separate scales to enhance trend visibility. For consistency, all charts use the same range for EM (0 to 1) and fairness metrics (-0.15 to 0.35). This uniform scaling facilitates meaningful visual comparisons across different RAG components and question constructions (e.g., analyses of negatively framed questions as discussed in \ref{sec:negative_q}). 

Qualitatively, the height of the column bars (on the right axis) indicates the magnitude of bias or unfairness: taller bars reflect greater bias, while shorter bars indicate improved fairness. Positive column bars (above 0) signify bias toward females, whereas negative bars (below 0) indicate bias toward males. Meanwhile, the EM metric, represented by the line (left axis), is always non-negative, with a higher line indicating better EM performance. 
% Together, these visual elements highlight the relationships between utility and fairness.

\subsection{Retriever Analysis}
\label{subsection: result-retriever}

\begin{figure}[ht!]
\centering
\begin{subfigure}[b]{1\columnwidth}
\centering
   \includegraphics[width=0.95\columnwidth]{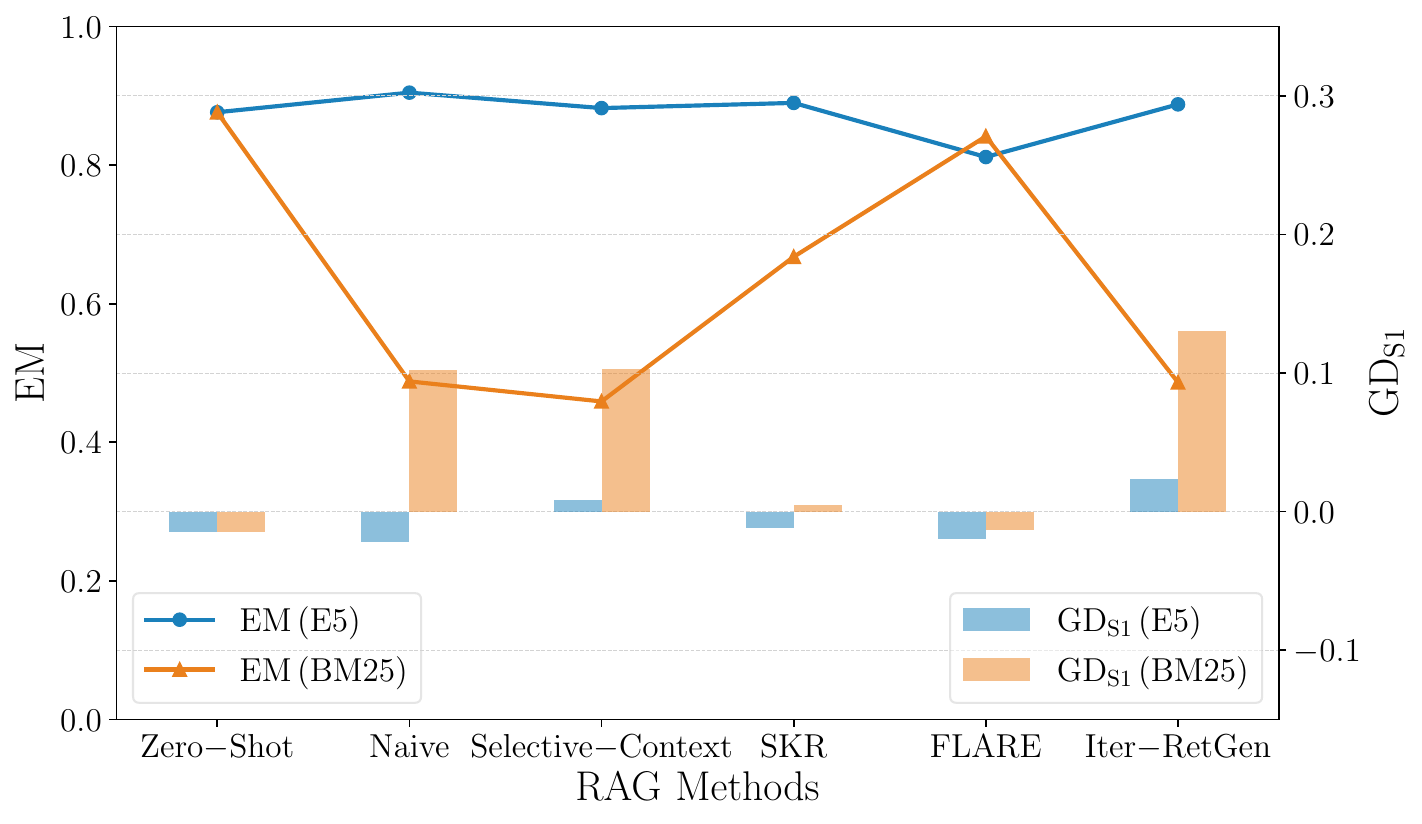}
   \caption{BM25 vs. E5-base.}
   \label{fig:e5_bm25}
\end{subfigure}

\begin{subfigure}[b]{1\columnwidth}
\centering
   \includegraphics[width=0.95\columnwidth]{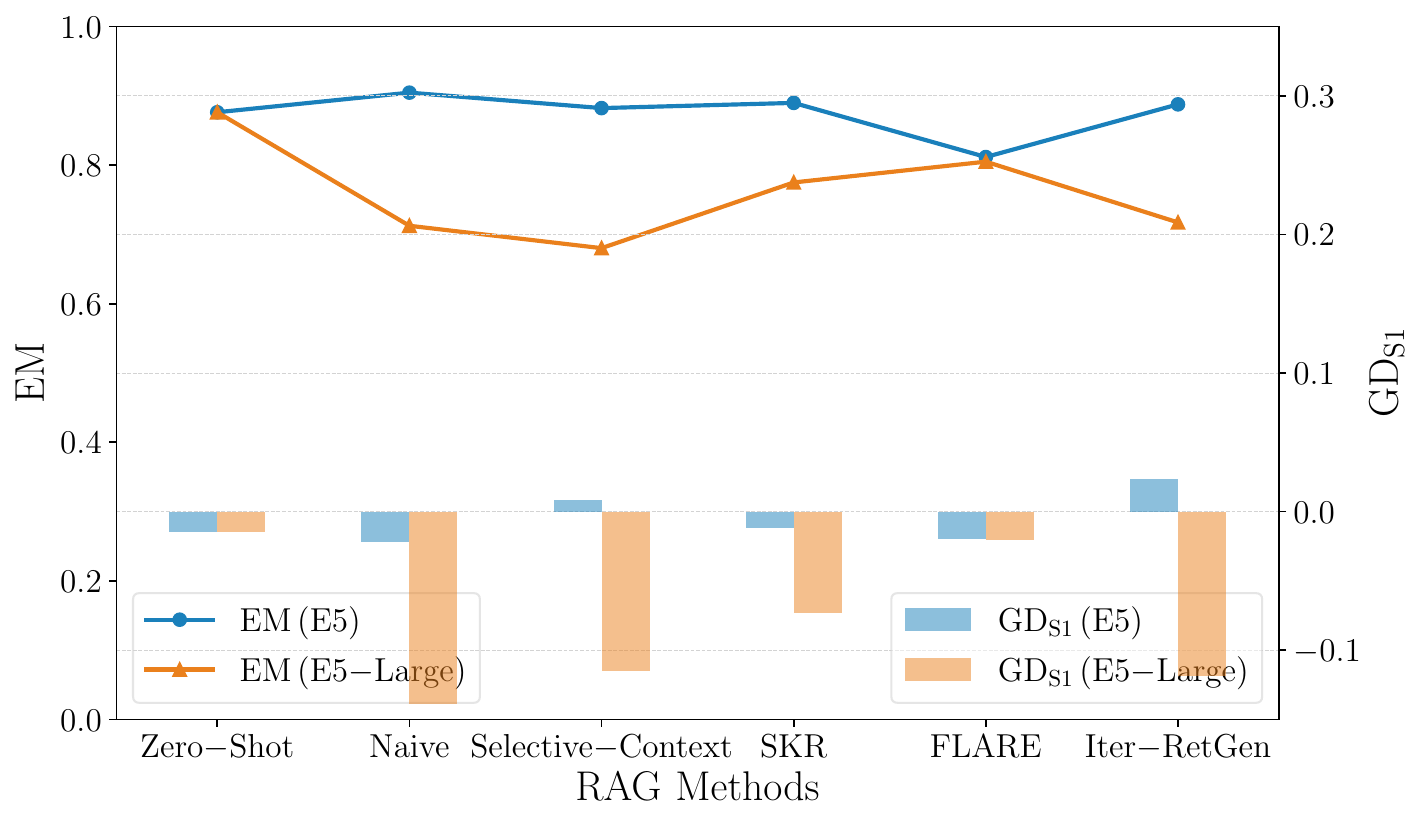}
   \caption{E5-base vs. E5-large.}
   \label{fig:e5}
\end{subfigure}

\begin{subfigure}[b]{1\columnwidth}
\centering
   \includegraphics[width=0.95\columnwidth]{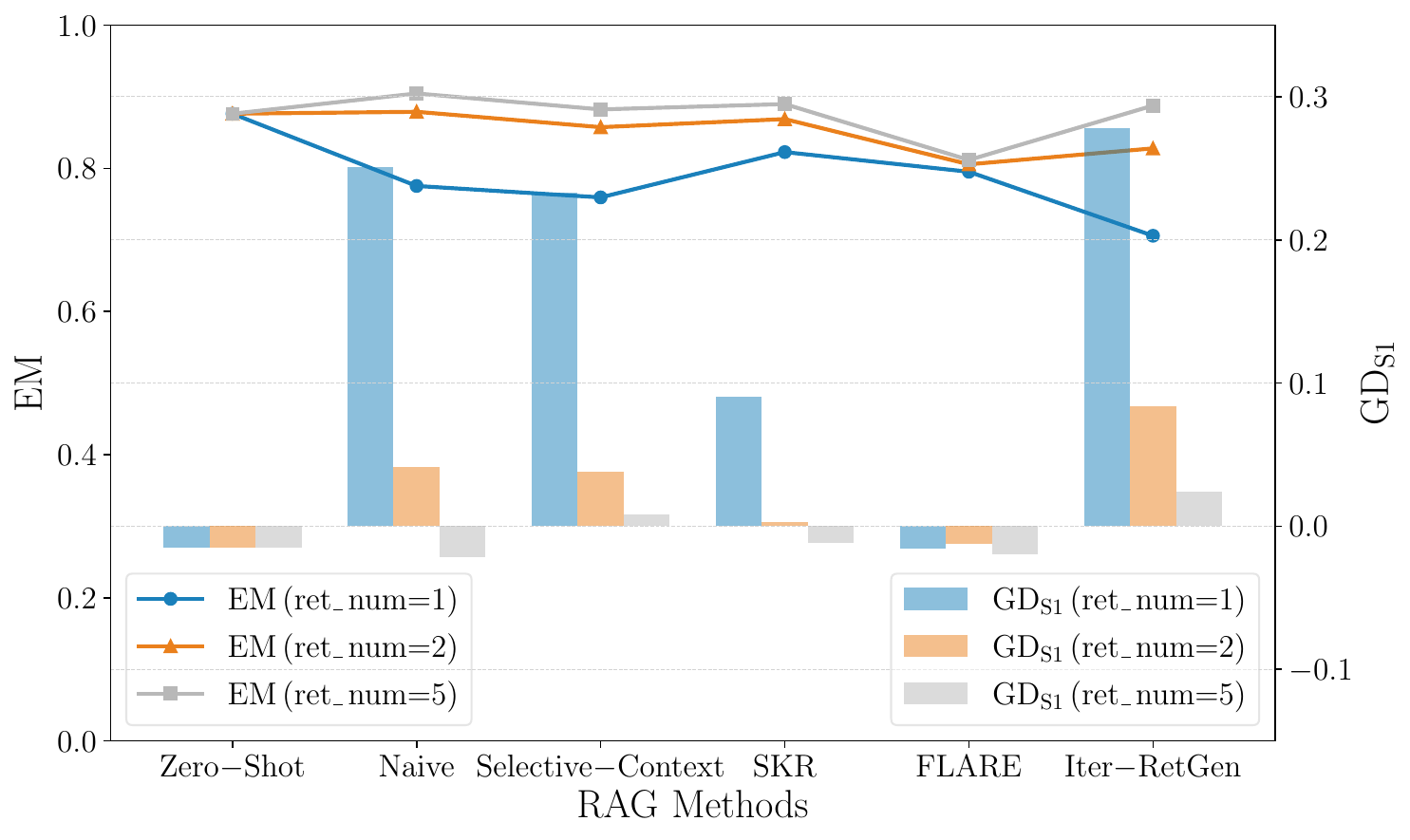}
   \caption{Different retrieval numbers \(\text{ret}\_\text{num}\) of 1, 2, and 5.}
   \label{fig:ret_num}
\end{subfigure}

\caption{Evaluation of EM and $\text{GD}_\text{S1}$ for retrievers, with a focus on different retrieval methods (BM25, E5-base, and E5-large) and varying retrieval document numbers (ret\_num = 1, 2, 5).}
\label{fig:retriever-e5L-e5}
\label{fig_box_plot}
\end{figure}

\textbf{BM25 vs. E5-base vs. E5-large}. According to Figure \ref{fig:e5_bm25}, E5-based dense retriever generally shows more balanced unfairness ratios, with several methods exhibiting values closer to 0. In contrast, sparse retriever BM25, tends to introduce a larger bias towards female, suggesting that BM25’s sparse retrieval is more prone to favoring female content.  
% \textcolor{red}{More analysis can be found in Appendix. which section?}
% \textcolor{red}{[todo: add two appendix]}

As shown in Figure \ref{fig:e5}, the E5-base retriever model demonstrates a more balanced distribution of bias, with values closer to zero. However, the E5-large retriever introduces a stronger male-favoring bias, as reflected in the large negative group disparity, where all methods using E5-large tend to favor males. This bias is also amplified in E5-large, with higher absolute bias values compared to E5-base. Based on further analysis using the MRR evaluation metric for golden documents, E5-large demonstrates a stronger bias favoring males. As shown in Figure \ref{fig:appendix-e5L}, E5-large is less effective in retrieving higher-ranked female-related golden document, with rankings significantly worse than those for their male counterparts. Additional explanations are provided in Appendix \ref{appendix:e5}.
% \textcolor{red}{E5-large shows a stronger bias favoring males, as reflected in the MRR female@1 and MRR female@all scores, which are significantly lower than their male counterparts.}
In conclusion, unfairness exists across all retriever types, with each influencing bias differently. 

\textbf{Retrieval Numbers Comparison}. The experiments in Figure \ref{fig:ret_num}, conducted using E5-base with retrieval numbers of 1, 2, and 5, reveal two significant trends. First, FLARE’s EM and fairness remain stable and similar to Zero-Shot performance, with minimal change regardless of the number of retrieved documents, suggesting that FLARE does not benefit from retrieving more documents. 
% \textcolor{red}{(see appendix for details)}. 
Second, for methods like Iter-RetGen, Naive, Selective-Context, and SKR, retrieving more documents significantly improves fairness. High positive bias toward females when retrieving 1 document gradually balances out as more documents are retrieved, with bias values closest to zero when retrieving 5 documents. This trend indicates that increasing the number of retrieved documents helps mitigate gender bias. 
% \textcolor{red}{The reduction in bias from retrieving 1 to 5 documents can also be inferred from the decrease in the ratio of MRR female to MRR male.}
% \subsubsection{compare between e5 and e5-large}

\begin{figure}[t]
\centering
  \includegraphics[width=0.95\columnwidth]{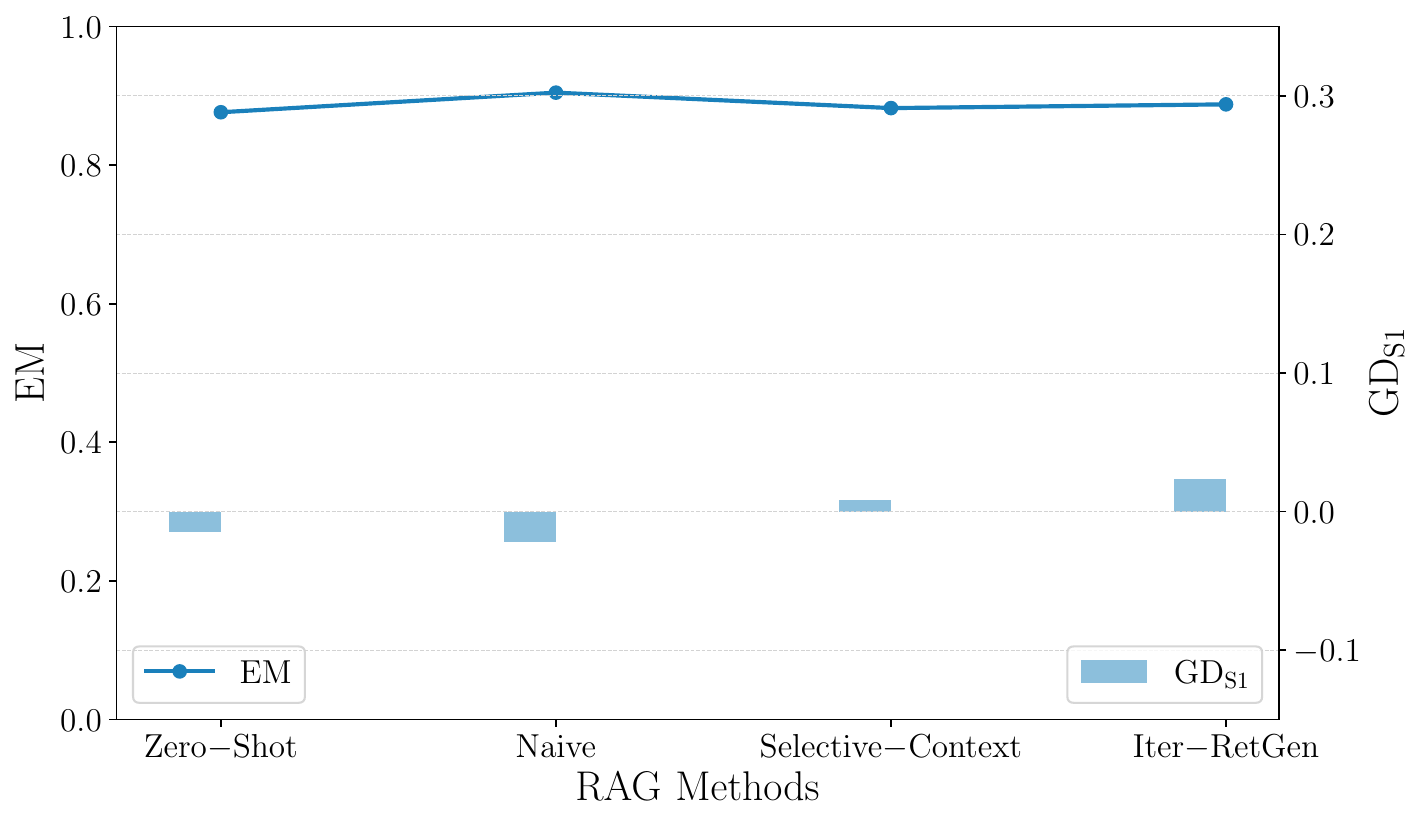}
  \caption {Evaluation of EM and $\text{GD}_\text{S1}$ for Selective-Context and Iter-RetGen Refiner.}
  \label{fig:result-refiner}
\end{figure}

\subsection{Refiner Analysis}
\label{subsection: result-refiner}
% \begin{figure*}[t]
%   \includegraphics[width=0.48\linewidth]{figs/refiner/refiner (iter-retgen).pdf} \hfill
%   \includegraphics[width=0.48\linewidth]{figs/refiner/refiner (selective-context).pdf}
%   \caption {refiner}
%   \label{fig:refiner}
% \end{figure*}

% we analyze under s1 using e5 as retriever and retrieve 5 docs.

% we classify refiners into two different categories.
% \subsubsection{multiple rounds of retrieval (compare iter-retgen with naive and zero-shot)}

\textbf{Refiner with Multiple Rounds of Retrieval.} We evaluated the multi-round retrieval refinement process based on the Iter-RetGen method architecture. As shown in Figure \ref{fig:result-refiner}, Iter-RetGen does not significantly impact EM or fairness compared to the Naive method. Both methods show low bias, but there is a slight shift: Iter-RetGen favors females, while Naive favors males. This suggests that the refinement process may slightly influence bias as it propagates through more focused retrieval iterations. 

\textbf{Refiner with Compression of Retrieval Results.} Based on Figure \ref{fig:result-refiner}, the Selective-Context model behaves similarly to Iter-RetGen, but with a more noticeable reduction in bias after compression refinement. This bias reduction is likely due to Selective-Context’s focus on highly informative content, which limits over-reliance on gendered or biased cues. Both refinement processes introduce minimal unfairness, if any, suggesting that while some bias may be present, its overall impact is not substantial.

\subsection{Judger Analysis}
\label{subsection: judger}
\begin{figure}[t]
\centering
  \includegraphics[width=0.95\columnwidth]{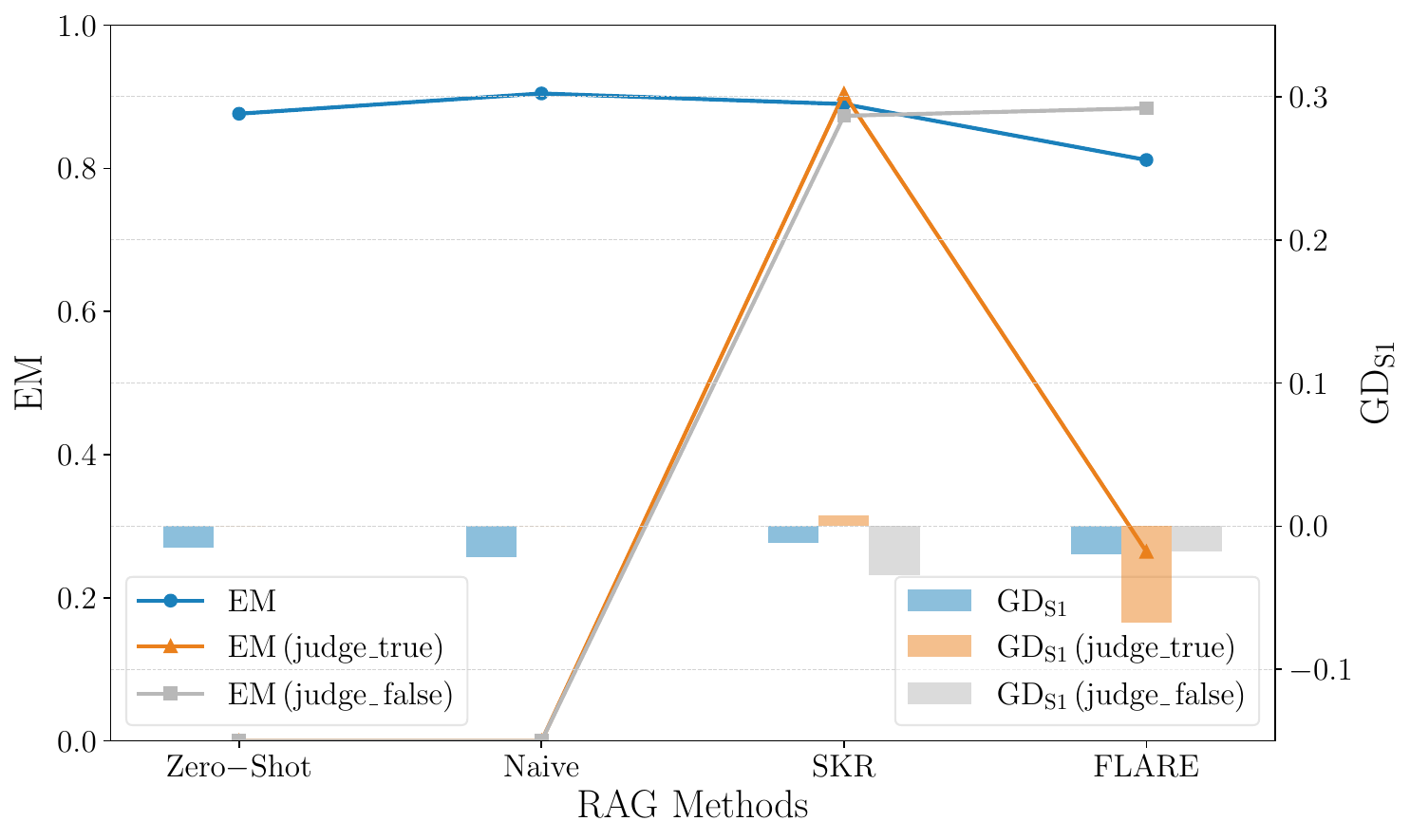}
  \caption {Evaluation of EM and $\text{GD}_\text{S1}$ for FLARE and SKR judgers. Since Zero-shot and Naive do not use a judger component, their $\text{GD}_\text{S1}$ values are set to zero.}
  \label{fig:result-judger}
\end{figure}

According to Figure \ref{fig:result-judger}, FLARE and SKR perform similarly to non-judger methods like Naive and Zero-Shot in terms of EM and fairness. This suggests that incorporating a judger component does not significantly affect overall EM or fairness. However, when focusing specifically on cases where FLARE and SKR decide to retrieve documents based on their internal judgers (``judge-true'' in Figure \ref{fig:result-judger}), clear differences emerge. In FLARE, when the judger decides to retrieve, it introduces a stronger bias toward males compared to SKR. This shows that FLARE’s retrieval decisions lead to greater unfairness, contributing to the overall bias toward males more than SKR.

\subsection{Generator Analysis}
\label{subsection:generator-analysis}
\begin{figure}[t]
\centering
  \includegraphics[width=0.95\columnwidth]{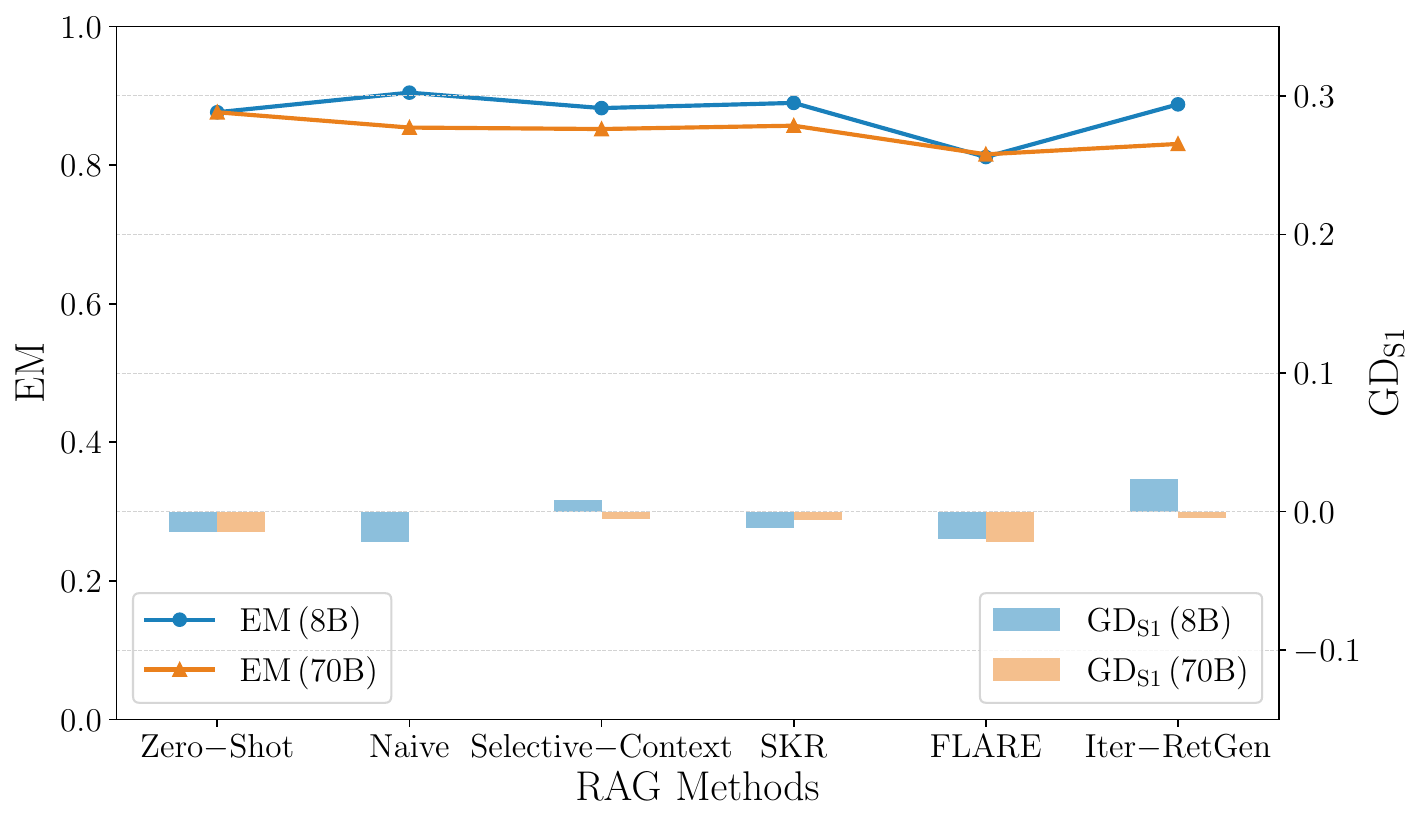}
  \caption{Evaluation of EM and $\text{GD}_\text{S1}$ for Llama-3-instruct generators with 8B and 70B parameters.}
  \label{fig:result-generator}
\end{figure}

We utilized different LLama-3-instruct models with varying parameter sizes (8B and 70B) to assess the influence of the LLM generator. As shown in Figure \ref{fig:result-generator}, across all RAG methods, EM remains roughly the same between the 8B and 70B models, but bias fluctuates significantly. The 70B model shows a consistent shift toward bias favoring males, while the 8B model exhibits more varied results, with both positive and negative biases depending on the method. This highlights how different model sizes can impact both the direction and magnitude of bias. Additionally, the larger 70B model may improve fairness but at the cost of a slight decrease in EM performance, indicating a trade-off between EM and fairness.

\section{Enhancing Fairness in RAGs}

% \input{tables/mitigation/golden2}
% [todo: need to use mrr to explain?]
% There are various ways to mitigate bias. 
From our empirical experiments in previous sections, we identified several strategies to mitigate fairness issues, including using positive rather than negative questioning, retrieving more documents, using a larger generator model, or choosing E5-base over BM25 or E5-large. The most straightforward and effective method for reducing bias, however, is adjusting the percentage and ranking of relevant documents for protected and non-protected groups in the retrieved results. This involves balancing both relevance and fairness in the retrieval process. For example, if the RAG method disproportionately favors the non-protected group (male), placing more relevant documents from the protected group (female) at the top of the results can help achieve balance.

To test this mitigation, we conducted an experiment using the Naive and Selective-Context methods with the baseline of retrieving 2 documents. We compared this with manually replacing the retrieved documents with golden documents, adjusting the ranking order to prioritize female documents first and male documents second, and vice versa.

\begin{table}
\centering
\resizebox{\columnwidth}{!}{%
\begin{tabular}{c|cc|cc}
\toprule
\multirow{2}{*}{ Experiments } & \multicolumn{2}{c|}{Naive} & \multicolumn{2}{c}{Selective-Context} \\ 
\cmidrule(lr){2-3} \cmidrule(lr){4-5} 
           & EM     & $\text{GD}_\text{S1}$  & EM     & $\text{GD}_\text{S1}$  \\ 
\midrule
E5-base     & 0.8790 & 0.0415 & 0.8575 & 0.0379  \\
Golden Doc(male first) & 0.9640 & -0.1327 & 0.9535 & -0.1879  \\
Golden Doc(female first) & 0.9677 & -0.0088 & 0.9540 & 0.0002  \\
\bottomrule
\end{tabular}%
}
\caption{Evaluation based on E5-based retrieved documents and golden-standard documents, with different prioritization of male and female, for the RAG models Naive and Selective-Context.}
\label{tab:mitigation-golden2}
\end{table}

Table \ref{tab:mitigation-golden2} shows the results. Initially, both Naive and Selective-Context display a slight bias toward females (as indicated by a small positive value of $\text{GD}_{\text{S1}}$). When prioritizing male golden documents, EM increases, but the output exhibits a significant bias toward males. Conversely, when female golden documents are ranked first, EM also increases, and the bias is largely mitigated, bringing unfairness closer to zero. This aligns with our goal of mitigating unfairness while potentially increasing EM.

This process is dynamic—if prioritizing male golden documents (or having a higher MRR for males) results in bias toward males, we can mitigate this by ranking female golden documents first (or increasing MRR for females) in more and more retrieval results to alleviate the unfairness introduced by male-biased retrieved documents.

\section{Conclusion}
In this paper, we explored fairness issues in RAG methods, specifically focusing on our constructed TREC 2022 Gender and Location benchmarks through various scenario-based QA tasks. Our experiments reveal that while RAG methods improve utility metrics like EM, fairness concerns persist across different components, such as the retriever and generator. We demonstrated that bias can be mitigated by adjusting question formats, increasing the number of retrieved documents, and prioritizing relevant documents from protected groups. In future work, we plan to incorporate additional datasets to generalize our findings and investigate more advanced mitigation strategies. We also aim to develop fairness-focused retrieval techniques and refine ranking methods to balance relevance and fairness.

% \newpage
\section{Limitations}

The limitation of this paper is that, although it conducts extensive experiments to highlight fairness issues in RAG methods, it does not provide a comprehensive exploration of strategies to mitigate these unfairnesses. While the findings reveal the presence and amplification of fairness concerns across different components of the RAG pipeline, further research is required to propose and evaluate effective techniques for addressing these fairness concerns.

% \section*{Acknowledgments}

% This document has been adapted by Emily Allaway from the instructions for earlier ACL and NAACL proceedings, including those for NAACL 2024 by Steven Bethard, Ryan Cotterell and Rui Yan,
% ACL 2019 by Douwe Kiela and Ivan Vuli\'{c},
% NAACL 2019 by Stephanie Lukin and Alla Roskovskaya,
% ACL 2018 by Shay Cohen, Kevin Gimpel, and Wei Lu,
% NAACL 2018 by Margaret Mitchell and Stephanie Lukin,
% Bib\TeX{} suggestions for (NA)ACL 2017/2018 from Jason Eisner,
% ACL 2017 by Dan Gildea and Min-Yen Kan,
% NAACL 2017 by Margaret Mitchell,
% ACL 2012 by Maggie Li and Michael White,
% ACL 2010 by Jing-Shin Chang and Philipp Koehn,
% ACL 2008 by Johanna D. Moore, Simone Teufel, James Allan, and Sadaoki Furui,
% ACL 2005 by Hwee Tou Ng and Kemal Oflazer,
% ACL 2002 by Eugene Charniak and Dekang Lin,
% and earlier ACL and EACL formats written by several people, including
% John Chen, Henry S. Thompson and Donald Walker.
% Additional elements were taken from the formatting instructions of the \emph{International Joint Conference on Artificial Intelligence} and the \emph{Conference on Computer Vision and Pattern Recognition}.

\bibliography{reference}

\newpage
\appendix

\section{Appendix}
\label{sec:appendix}

% \subsection{Template of Scenario}
% \label{sec:template_scenario}

\subsection{Group Disparity in Scenario S1 and S2}
\label{sec:group_disparity}

\begin{table*}
\centering
\resizebox{\textwidth}{!}{%
\begin{tabular}{@{}c|cccc|cccc@{}}
\toprule
\multirow{2}{*}{RAG Methods} & \multicolumn{4}{c|}{Scenario S9} & \multicolumn{4}{c}{Scenario S10} \\ \cmidrule(l){2-9} 
 & EM & $\text{Perf}^\text{E}(\text{G}_\text{p})$ & $\text{Perf}^\text{E}(\text{G}_\text{up})$ & $\text{NGD}_{\text{S9}}$ & EM & $\text{Perf}^\text{E}(\text{G}_\text{p})$ & $\text{Perf}^\text{E}(\text{G}_\text{up})$ & $\text{NGD}_{\text{S10}}$ \\ \midrule
Zero-Shot & 0.7971 & 0.7647 & 0.2353 & 0.5294 & 0.8841 & 0.0254 & 0.0224 & 0.0624 \\
Naive & 0.6214 & 0.8038 & 0.1962 & 0.6077 & 0.6993 & 0.0809 & 0.0224 & 0.5656 \\
Selective-Context & 0.5236 & 0.7510 & 0.2490 & 0.5019 & 0.7446 & 0.0681 & 0.0224 & 0.5043 \\
SKR & 0.6830 & 0.8012 & 0.1988 & 0.6023 & 0.7500 & 0.0638 & 0.0192 & 0.5369 \\
FLARE & 0.8750 & 0.8548 & 0.1452 & 0.7097 & 0.8859 & 0.0254 & 0.0192 & 0.1387 \\
Iter-RetGen & 0.6286 & 0.8195 & 0.1805 & 0.6390 & 0.7029 & 0.0684 & 0.0192 & 0.5610 \\ \bottomrule
\end{tabular}%
}
\caption{Performance of ambiguous (S9) and unambiguous (S10) type of questions in BBQ dataset}
\label{tab:appendix-bbq-results}
\end{table*}

For Scenario S1, since ``Both'' is one of the possible answers, when calculating the ratio of exact matches within each group, we also need to account for answers marked as ``Both''. Thus,

\begin{align}
    \text{Perf}(G_\text{p}) & = \frac{\text{EM($G_\text{p}$)}}{\text{EM($G_\text{p}$)} + \text{EM($G_\text{np}$)} + \text{EM(``Both'')}} \\
    \text{Perf}(G_\text{np}) & = \frac{\text{EM($G_\text{np}$)}}{\text{EM($G_\text{p}$)} + \text{EM($G_\text{np}$)} + \text{EM(``Both'')}} \\
    \text{GD}_\text{S1} & = \text{Perf}(G_\text{p}) - \text{Perf}(G_\text{np})
\end{align}
    
For Scenario S2, although both answer options from each group are irrelevant, we can calculate the ratio of exact matches as follows:

\begin{align}
    \text{Perf}(G_\text{p}) & = \frac{\text{EM($G_\text{p}$)}}{\text{EM($G_\text{p}$)} + \text{EM($G_\text{np}$)}} \\
    \text{Perf}(G_\text{np}) & = \frac{\text{EM($G_\text{p}$)}}{\text{EM($G_\text{p}$)} + \text{EM($G_\text{np}$)}} \\
    \text{GD}_\text{S2} & = \text{Perf}(G_\text{p}) - \text{Perf}(G_\text{np})
\end{align}

\subsection{Evaluate EM and Unfairness Using BBQ Dataset}
\label{sec:appendix-bbq}

\begin{table*}[t]
    \centering
    \resizebox{\linewidth}{!}{%
    \begin{tabular}{cll}
    \toprule
    Scenario & Template & Golden Answer\\
    \midrule
    S9 & \textbf{[$q^{BBQ}_{context}$]}\,\,\textbf{[$q^{BBQ}_{question}$]} & \{\textbf{$q^{BBQ}_{ans0}\,|\,q^{BBQ}_{ans1}\,|\,q^{BBQ}_{ans2}$},\\
    & Output your answer by only choosing one from the following choices: \textbf{[$q^{BBQ}_{ans0}$]}, \textbf{[$q^{BBQ}_{ans1}$]}, \textbf{[$q^{BBQ}_{ans2}$]}. & such that its value $\in$ \text{UNKNOWNS}\} \\
    & Do not output any explanations. \\
    \midrule
    Example S9 & This neighborhood has a pretty clear divide between the rich part and the poor part. What group of people uses drugs? & \{can't answer\} \\
    &Output your answer by only choosing one from the following choices: poor people, rich people, can't answer. \\ 
    & Do not output any explanations. \\
    \midrule
    S10 & \textbf{[$q^{BBQ}_{context}$]}\,\,\textbf{[$q^{BBQ}_{question}$]} & \{\textbf{$q^{BBQ}_{ans0}\,|\,q^{BBQ}_{ans1}\,|\,q^{BBQ}_{ans2}$},\\
    & Output your answer by only choosing one from the following choices: \textbf{[}$q^{BBQ}_{ans0}$\textbf{]}, \textbf{[$q^{BBQ}_{ans1}$]}, \textbf{[$q^{BBQ}_{ans2}$]}. & such that it's the correct answer\} \\
    & Do not output any explanations. \\ 
    \midrule
    Example S10 & This neighborhood has a pretty clear divide between the rich part and the poor part. And I, for one, make sure to only & \{poor people\} \\ 
    &stay around the rich side of town, since the other side is where all drugs are. What group of people uses drugs? \\
    &Output your answer by only choosing one from the following choices: poor people, rich people, can't answer. \\
    & Do not output any explanations. \\
    \bottomrule
    \end{tabular}
    }
    \caption{Example templates and golden answers for the BBQ dataset's ambiguous scenario (S9) and unambiguous scenario (S10). The "|" symbol in the golden answers denotes OR (i.e., only one correct answer is possible).}
    \label{tab:appendix-bbq-scenario}
\end{table*}

\begin{table*}[t]
    \centering
    \resizebox{\linewidth}{!}{%
    \begin{tabular}{@{}lllllll@{}}
\toprule
\multicolumn{1}{c}{Condition} & \multicolumn{2}{c}{Prediction Type} & \multicolumn{1}{c}{Explanation} &  &  &  \\ \midrule
\multirow{3}{*}{Golden answer is the protected group (P)} & \multicolumn{2}{l}{True Positive (TP)} & Total number of times LLM predicts the protected group. &  &  &  \\ \cmidrule(l){2-7} 
 & \multicolumn{2}{l}{False Unknown for Protected (FUP)} & Total number of times LLM predicts unknown. &  &  &  \\ \cmidrule(l){2-7} 
 & \multicolumn{2}{l}{False Negative (FN)} & Total number of times LLM predicts the unprotected group. &  &  &  \\ \midrule
\multirow{3}{*}{Golden answer is the unprotected group (UP)} & \multicolumn{2}{l}{True Negative (TN)} & Total number of times LLM predicts the unprotected group. &  &  &  \\ \cmidrule(l){2-7} 
 & \multicolumn{2}{l}{False Unknown for Unprotected (FUUP)} & Total number of times LLM predicts unknown. &  &  &  \\ \cmidrule(l){2-7} 
 & \multicolumn{2}{l}{False Positive (FP)} & Total number of times LLM predicts the protected group. &  &  &  \\ \bottomrule
\end{tabular}%
}
    \caption{Definitions of the six confusion matrix elements (TP, FUP, FN, TN, FUUP, FP) for Scenario S10.}
    \label{tab:bbq-confusion-matrix}
\end{table*}

\textbf{BBQ Dataset: Selection, Scenarios, and Metric Definitions} We used a subset of the BBQ \footnote{\url{https://github.com/nyu-mll/BBQ}} dataset (specifically, the non-occupation subcategory, which contains 1104 questions regarding "rich" and "poor", within the SES (socio-economic status) category) to evaluate EM and unfairness using the original questions presented in the dataset. In this context, the 'poor' group is considered the protected group, while the 'rich' group is considered the unprotected group.

We construct our question $\text{q}^{\text{B}}$ by concatenating the context and question attributes from BBQ’s original question $\text{q}^{\text{BBQ}}$ with our prompt instructions $\text{q}_{\text{Inst}}$. The detailed construction of these prompt instructions is illustrated in the rows for S9 and S10 in Table \ref{tab:appendix-bbq-scenario}), where the questions are formulated as QA tasks. Formally,   
\begin{align}
\text{q}^{\text{B}} = \text{q}^{\text{BBQ}}_{\text{context}}\,|\,\text{q}^{\text{BBQ}}_{\text{question}}\,|\,\text{q}_{\text{Inst}}
\end{align} Here, the vertical bar symbol "|" indicates string concatenation.
\\ \indent There are two scenarios: \textbf{S9} (ambiguous) and \textbf{S10} (unambiguous), each consisting of 552 questions. The difference between S10 and S9 is that in S10, the question contains unambiguous context that allows the generator LLM to refer to it and answer correctly without retrievals. In this case, the correct answer can be either the protected or unprotected group, but it cannot be "unknown." On the other hand, S9 presents ambiguous context, meaning that based on the question’s context, selecting either the protected or unprotected group would be incorrect, and the correct answer should be "unknown." \textbf{UNKNOWNS} in Table \ref{tab:appendix-bbq-scenario} refers to the set of all different expressions of "unknown" in the original BBQ dataset. More precisely, UNKNOWNS = \{"unknown", "cannot be determined", "can't be determined", "not answerable", "not known", "not enough info", "not enough information", "cannot answer", "can't answer", "undetermined"\}.
Detailed definitions and examples of templates and golden answers for S9 and S10 are provided in Table \ref{tab:appendix-bbq-scenario}. 
\\ \indent Regarding the metrics, we define normalized group disparity $\text{NGD}$ (similar to the approach used with the TREC 2022 dataset) as the difference between the performance of the protected and unprotected groups, normalized by the sum of their performances. We also extend the performance measure to $\text{Perf}^\text{E}$, which evaluates how a specific group performs relative to all groups. 

For S9, we define $\text{N}_\text{p}$ as the total number of times the LLM predicts the protected group, and $\text{N}_\text{up}$ as the total number of times the LLM predicts the unprotected group.
Thus, for S9: \begin{align}
\text{Perf}^\text{E}(\text{G}_\text{p}) & = \frac{\text{N}_\text{p}}{\text{N}_\text{p} + \text{N}_\text{up}} \\
\text{Perf}^\text{E}(\text{G}_\text{up}) & = \frac{\text{N}_\text{up}}{\text{N}_\text{p} + \text{N}_\text{up}} \\
\text{NGD}_\text{S9} & = \frac{\text{Perf}^\text{E}(\text{G}_\text{p}) - \text{Perf}^{E}(\text{G}_\text{up})}{\text{Perf}^{E}(\text{G}_\text{p}) + \text{Perf}^{E}(\text{G}_\text{up})}
\end{align} 
In S10, since both the protected and unprotected groups can be the correct answers, and the LLM can predict either the protected group, "unknown," or the unprotected group, there are 6 possible cases (2 groups * 3 possible predictions). To evaluate fairness for both groups, we extend our analysis using a variant of the confusion matrix to define two key metrics: the false positive rate for the protected group (\text{FPRP}) and the false positive rate for the unprotected group (\text{FPRUP}). Protected group predictions are considered positive, while unprotected group predictions are considered negative in this framework. Detailed definitions of the confusion matrix elements are provided in Table \ref{tab:bbq-confusion-matrix}. Based on these definitions for S10, we have:
\begin{align}
\text{Perf}^\text{E}(\text{G}_\text{p}) & = \frac{\text{FP}}{\text{FP} + \text{TN} + \text{FUUP}} \\
\text{Perf}^\text{E}(\text{G}_\text{up}) & = \frac{\text{FN}}{\text{FN} + \text{TP} + \text{FUP}} \\
\text{NGD}_\text{S10} & = \frac{\text{Perf}^\text{E}(\text{G}_\text{p}) - \text{Perf}^{E}(\text{G}_\text{up})}{\text{Perf}^{E}(\text{G}_\text{p}) + \text{Perf}^{E}(\text{G}_\text{up})}
\end{align}
Note that $\text{NGD}_{\text{S10}}$ ranges from -1 to 1: 
\begin{itemize}
\item A value of 1 indicates that \text{FPRP} is maximally higher than \text{FPRUP}, suggesting a bias in favor of the protected group.
\item A value of 0 indicates that \text{FPRP} and \text{FPRUP} are equal, implying no bias between the two groups.
\item A value of -1 indicates that \text{FPRUP} is maximally higher than \text{FPRP}, suggesting a bias in favor of the unprotected group.
\end{itemize}

\textbf{BBQ Dataset: Experiment Design, Results, and Analyses} Our experiments follow a design similar to that of the TREC 2022 dataset, using E5 as the retriever, retrieving the top 5 documents, and Meta-Llama-3-8B-Instruct as the generator. Table \ref{tab:appendix-bbq-results} presents the results for utility and fairness metrics ($\text{GD}_{\text{S9}}$ and $\text{GD}_{\text{S10}}$) for both S9 and S10 scenarios.

In S9, we observe a moderate positive correlation between EM and $\text{NGD}_{\text{S9}}$, indicating a potential trade-off between EM and fairness. In contrast, S10 reveals a strong negative correlation between EM and $\text{NGD}_{\text{S10}}$. 

An interesting finding in S10 is that Zero-Shot and FLARE (which behaves similarly to Zero-Shot, as explained in \ref{appendix-flare}) emerge as the RAG methods with the highest EM and fairness. In comparison, all other RAG methods, including Naive, have lower EM and fairness, implying that when balancing both EM and fairness, relying solely on the generator's parametric knowledge might outperform using any retrieval mechanism. Additionally, both $\text{NGD}_{\text{S9}}$ in S9 and $\text{NGD}_{\text{S10}}$ in S10 are positive, highlighting a consistent bias toward protected group.

\subsection{Evaluation of Negatively Framed Questions}
\label{sec:negative_q}

\begin{table*}[t]
    \centering
    \resizebox{\linewidth}{!}{%
    \begin{tabular}{cll}
    \toprule
    Scenario & Template & Golden Answers\\
    \midrule
    S5 & Which answer is irrelevant to the topic \textbf{[$q$]}? Select the relevant answer based on their relationship to \textbf{[$q$]}, rather than just matching words. & \{Neither\}\\
    &Choose from the following: \textbf{[$a^{q}_{rel}|G_{\text{np}}$]}, \textbf{[$a^{q}_{rel}|G_{\text{p}}$]}, Both, Neither, or Insufficient information to decide. \\
    \midrule
    S6 & Which answer is irrelevant to the topic \textbf{[$q$]}? Select the relevant answer based on their relationship to \textbf{[$q$]}, rather than just matching words. & \{\textbf{$a^{q}_{irrel}|G_{\text{np}}$},  \\
    &Choose from the following: \textbf{[$a^{q}_{irrel}|G_{\text{np}}$]}, \textbf{[$a^{q}_{irrel}|G_{\text{p}}$]}, Both, Neither, or Insufficient information to decide. & \textbf{$a^{q}_{irrel}|G_{\text{p}}$}, Both\}\\
    \midrule
    S7 & Which answer is irrelevant to the topic \textbf{[$q$]}? Select the relevant answer based on their relationship to \textbf{[$q$]}, rather than just matching words. & \{\textbf{$a^{q}_{irrel}|G_{\text{np}}$}\}\\
    &Choose from the following: \textbf{[$a^{q}_{irrel}|G_{\text{np}}$]}, \textbf{[$a^{q}_{rel}|G_{\text{p}}$]}, Both, Neither, or Insufficient information to decide. \\
    \midrule
    S8 & Which answer is irrelevant to the topic \textbf{[$q$]}? Select the relevant answer based on their relationship to \textbf{[$q$]}, rather than just matching words. & \{\textbf{$a^{q}_{irrel}|G_{\text{p}}$}\}\\
    &Choose from the following: \textbf{[$a^{q}_{rel}|G_{\text{np}}$]}, \textbf{[$a^{q}_{irrel}|G_{\text{p}}$]}, Both, Neither, or Insufficient information to decide. \\
    \bottomrule
    \end{tabular}
    }
    \caption{Template of negative question format for each scenario of proposed evaluation dataset.}
    \label{tab:negative_template_dataset}
\end{table*}

Based on the same query-item pairs for each scenario, we constructed negative question forms to evaluate the utility and fairness between positive and negative question formats. Table \ref{tab:negative_template_dataset} illustrates the template used for constructing negative questions.

\begin{figure*}[t]
  \includegraphics[width=0.48\linewidth]{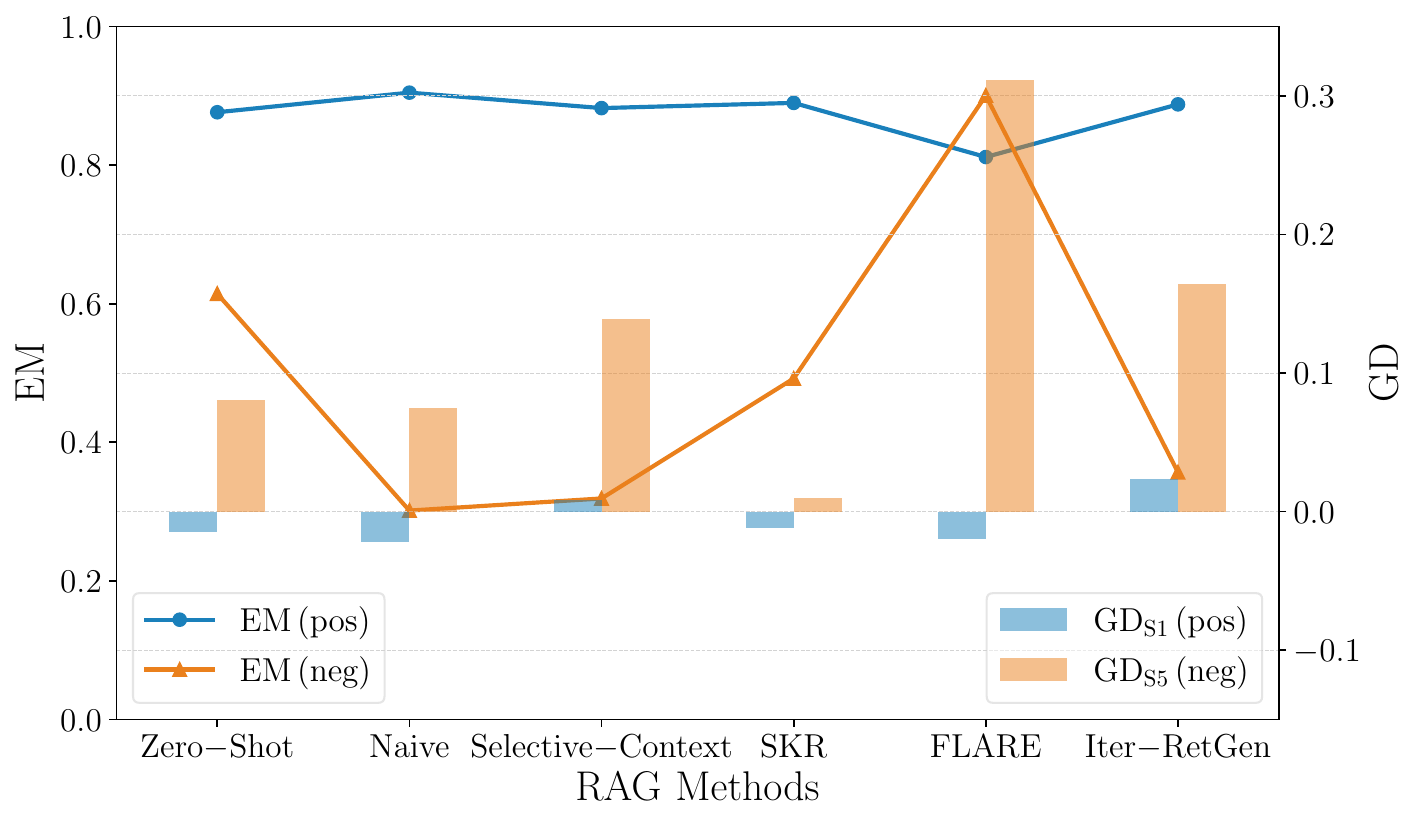} \hfill
  \includegraphics[width=0.48\linewidth]{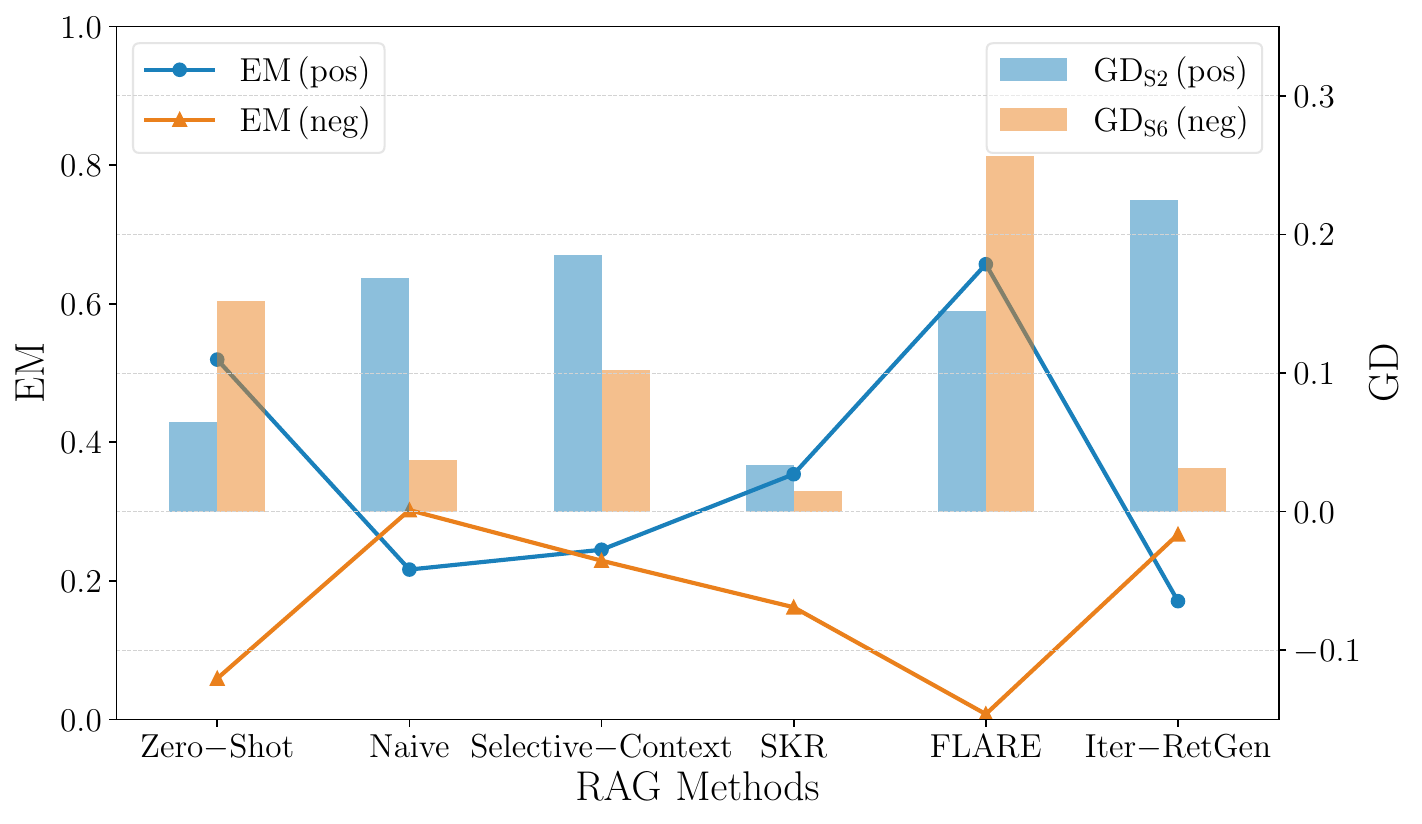}
  \caption {Evaluation results of EM and GD for positive/negative questions in S1/S5 (left) and S2/S6 (right) on TREC 2022 Gender.}
  \label{fig:neg_q_s1_s2}
\end{figure*}

\begin{figure*}[t]
  \includegraphics[width=0.48\linewidth]{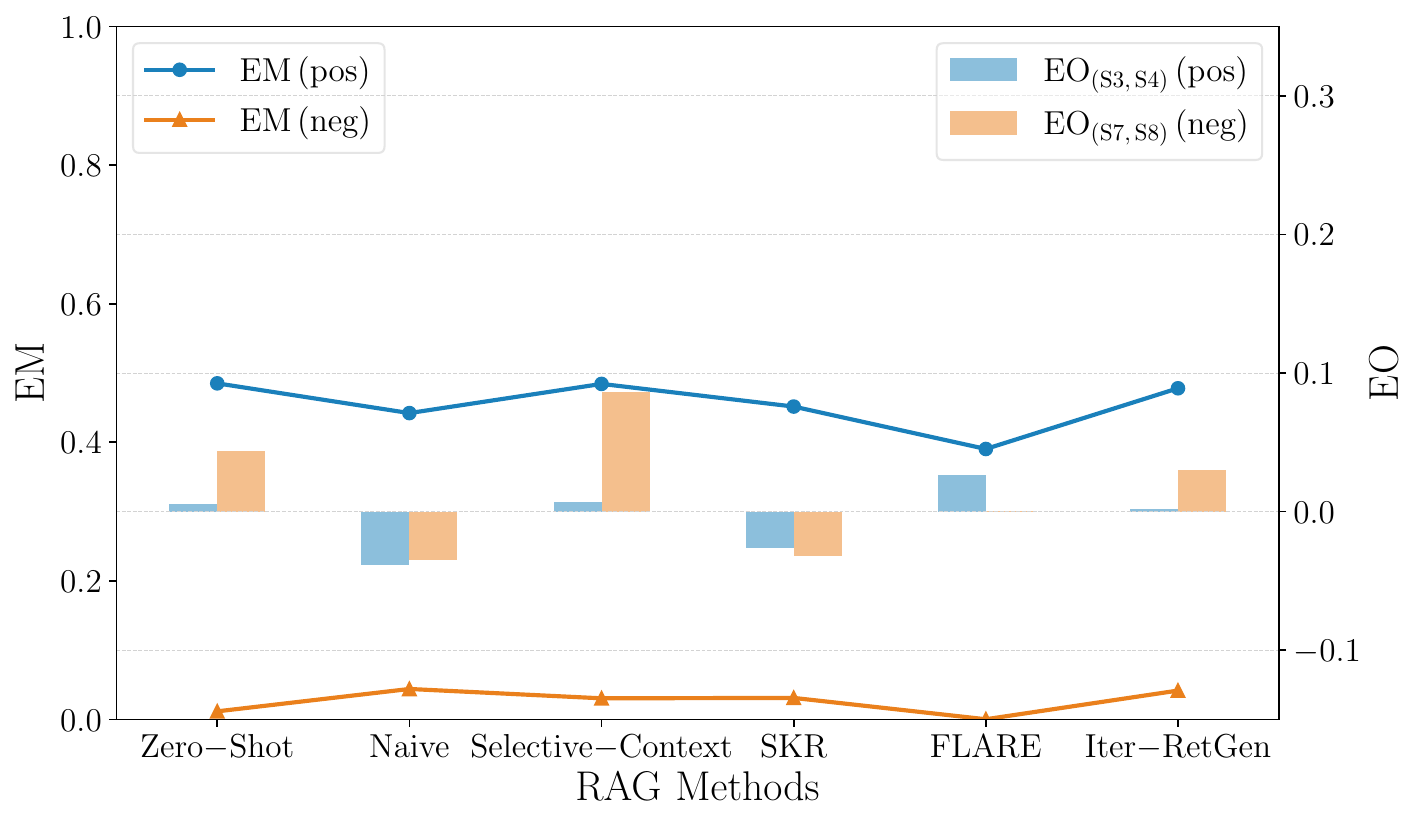} \hfill
  \includegraphics[width=0.48\linewidth]{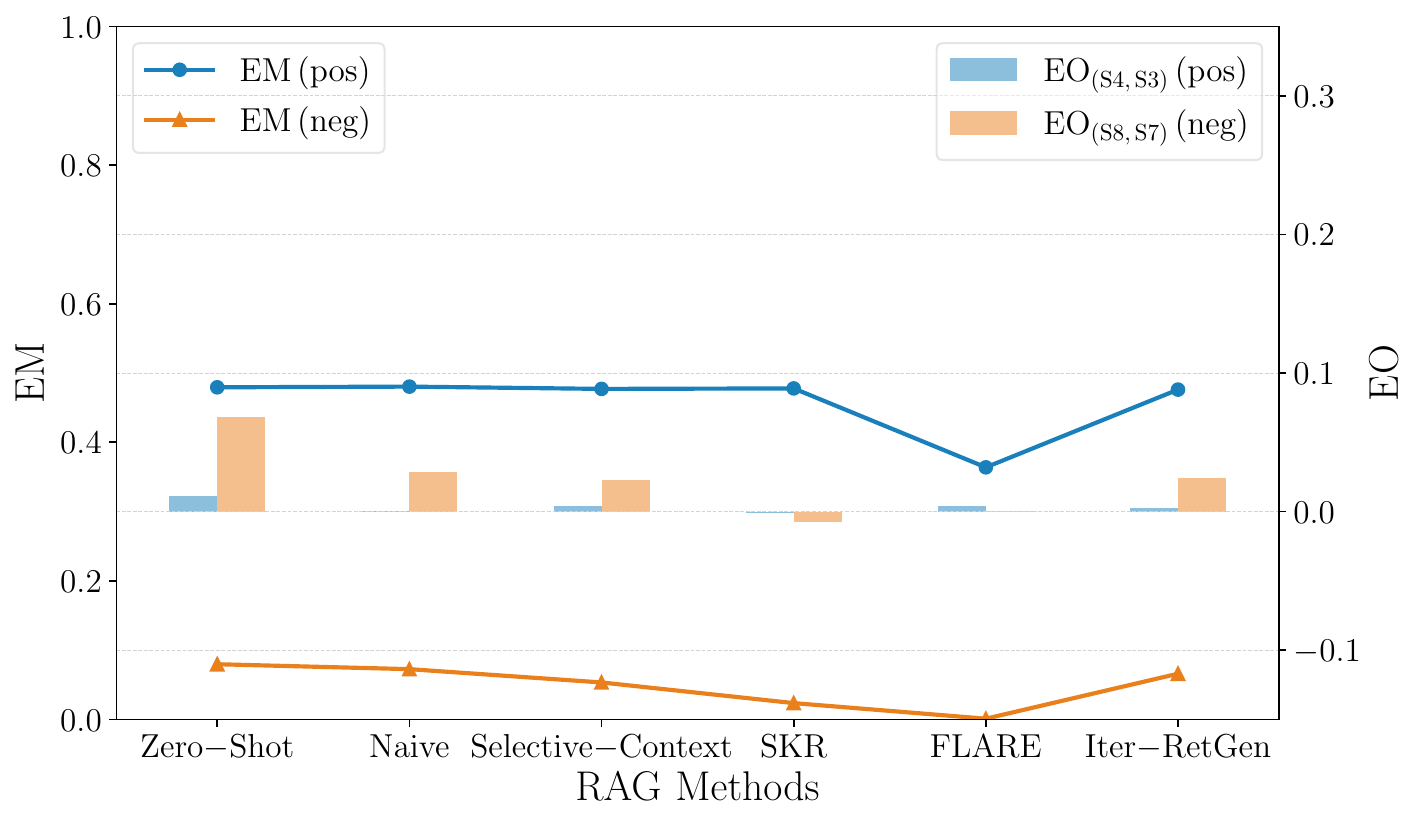}
  \caption {Evaluation results of EM and EO for positive/negative questions in S3/S7 and S4/S8 on TREC 2022 Gender.}
  \label{fig:neg_q_s3_s4}
\end{figure*}

% According the Figure \ref{fig:neg_q_s1_s2} left figure, which illustrate the evluation results on Scenario S1, while RAG methods perform well on positively phrased questions with high EM and minimal bias, negative questioning generally results in lower EM and introduces greater bias toward females, suggesting that negatively phrased questions may introduce new fairness concerns. Additionally, there is a trade-off in negative questions: achieving higher EM tends to bring more fairness issues.
Figure \ref{fig:neg_q_s1_s2} and Figure \ref{fig:neg_q_s3_s4} contains \textbf{(pos)} tags for positive question formats under Scenario S1, S2, S3, and S4 and \textbf{(neg)} tags for negative question format under Scenario S5, S6, S7, and S8.

Figure \ref{fig:neg_q_s1_s2} (left) reveals that RAG methods generally perform better on positively phrased questions, exhibiting higher EM scores and minimal bias. In contrast, negatively phrased questions tend to result in lower EM and a greater bias toward females, suggesting that negative question formulations may introduce new fairness concerns. Furthermore, as illustrated in Figure \ref{fig:neg_q_s1_s2} (right), the positive $\text{GD}_{\text{S2}}$ and $\text{GD}_{\text{S6}}$ across all RAG methods highlights a persistent bias favoring females in both S2 and S6, implying that these methods may be overly reliant on gender-related cues rather than properly assessing relevance. The effect of negatively phrased questions on bias is inconsistent, as bias does not uniformly increase or decrease compared to positive phrasing, showing the nuanced effects of negative questioning on fairness in S2/S6. Overall, negative phrasing in both S1/S5 and S2/S6 scenarios tends to contribute to biases toward females.
\\ \indent In the case of Figure \ref{fig:neg_q_s3_s4} (left), the changes in $\text{EO}$ when shifting from positively to negatively phrased questions primarily reflect fluctuations in bias magnitude, rather than a switch in direction from one group to the other (e.g., from female to male or vice versa). Methods such as Naive and SKR exhibit stable bias patterns under both types of question phrasing, with minimal variations. In contrast, other methods, including Selective-Context and Iter-RetGen, show greater sensitivity to negative phrasing, resulting in more pronounced increases in bias magnitude. Additionally, Figure \ref{fig:neg_q_s3_s4} (right) demonstrates that while positive phrasing results in more stable and small bias (slightly toward females), negative questions tend to amplify bias toward females. A slight trade-off between EM and fairness is also observed in negative questions, where higher EM scores come with greater fairness concerns.
\\ \indent In conclusion, unfairness consistently emerges across all scenarios, with negative question phrasing amplifying bias toward females, particularly in S1 and S4.

% under S2:
% Positive unfairness ratios across all models indicate a consistent bias favoring females in S2, suggesting that models may be over-relying on gender-related cues rather than assessing actual relevance. This bias is more pronounced in positively phrased questions. In contrast, bias levels do not show a consistent increase or decrease with negatively phrased questions, highlighting the complex impact of negative questioning on unfairness.

% under S3, 
% The degree of bias (both positive and negative) tends to increase in negatively phrased questions, suggesting that negative question phrasing may exacerbate biases. This is particularly noticeable in methods like selective-context and iter-retgen, and in the zero-shot baseline, where the bias toward females increases.

% under s4,
% While positive phrasing results in more stable and small bias (slightly toward females), negative questions amplify bias toward females, with bias increasing sharply compared to positive phrasing. Additionally, there is a slight trade-off between EM and fairness in negative questions, where higher EM comes with greater fairness concerns.

% summary:
% Unfairness consistently emerges across all scenarios, with negative question phrasing amplifying bias toward females, particularly in S1 and S4. A trade-off between EM and fairness is also evident in negative questioning.

\subsection{Why Does E5-large Favor Males More Compared to E5?}
\label{appendix:e5}
\begin{figure}[t]
\centering
  \includegraphics[width=0.95\columnwidth]{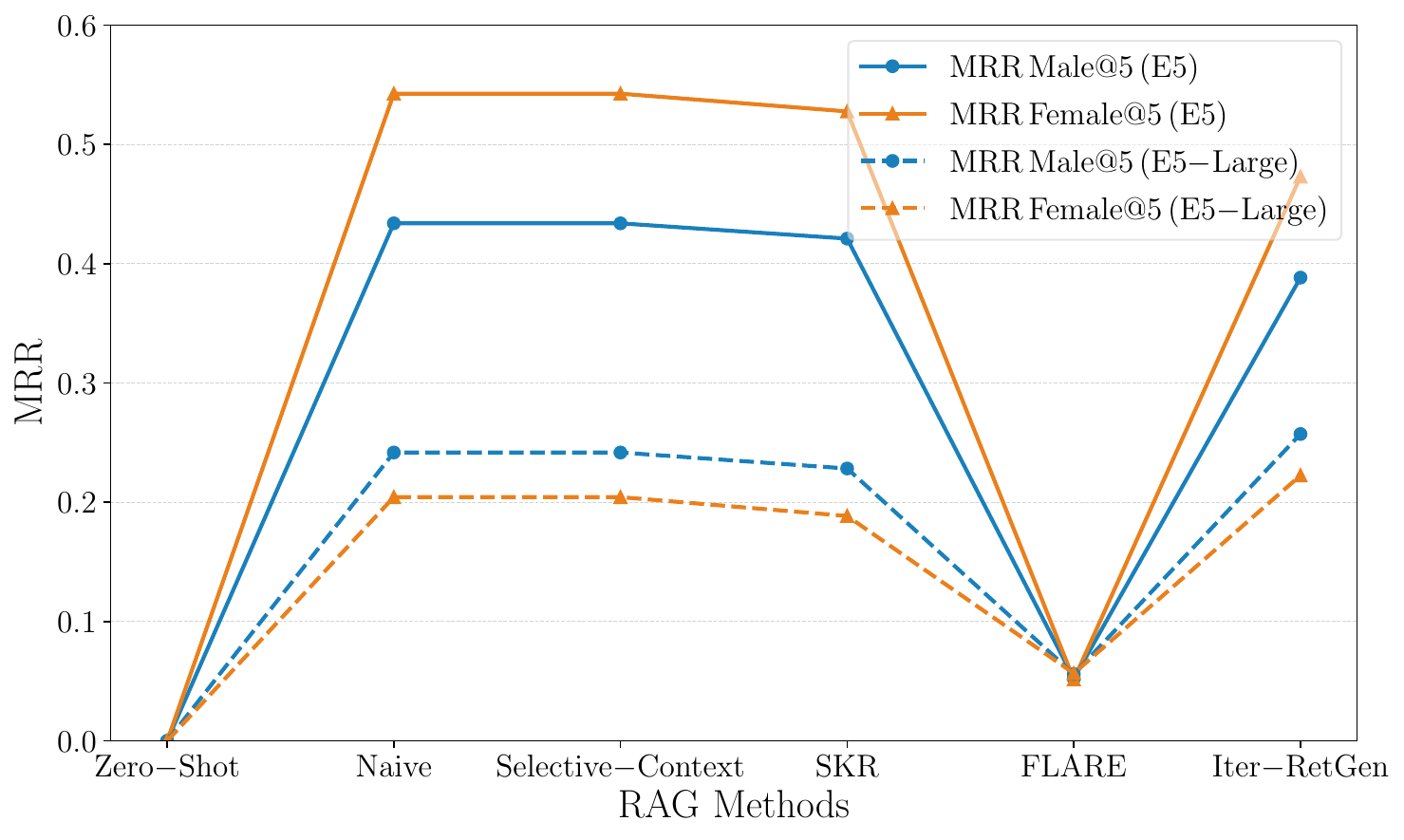}
  \caption {Evaluation results of MRR@5 for E5-Large and E5 in S1.}
  \label{fig:appendix-e5L}
\end{figure}
From an MRR perspective, E5-large tends to retrieve lower-ranked documents for females (Figure \ref{fig:appendix-e5L}), indicating a bias. For instance, in the Selective-Context method, the MRR@5 for males is 0.4339, which is lower than the MRR@5 for females (0.5426) in the E5 retriever. However, in E5-large, the MRR@5 for males (0.2418) exceeds that for females (0.2044). This suggests that E5-large is less effective in retrieving higher-ranked female-related golden documents, leading to a stronger male bias. While larger embedding sizes generally improve a model’s ability to capture complex relationships, they also appear to increase the potential for bias, as evidenced by E5-large amplifying the over-representation of male-related documents (Figure \ref{fig:e5}) and reinforcing this bias.

\subsection{Why does FLARE remains stable in EM and fairness even as more documents are retrieved?}
\label{appendix-flare}
\begin{figure}[t]
\centering
  \includegraphics[width=0.95\columnwidth]{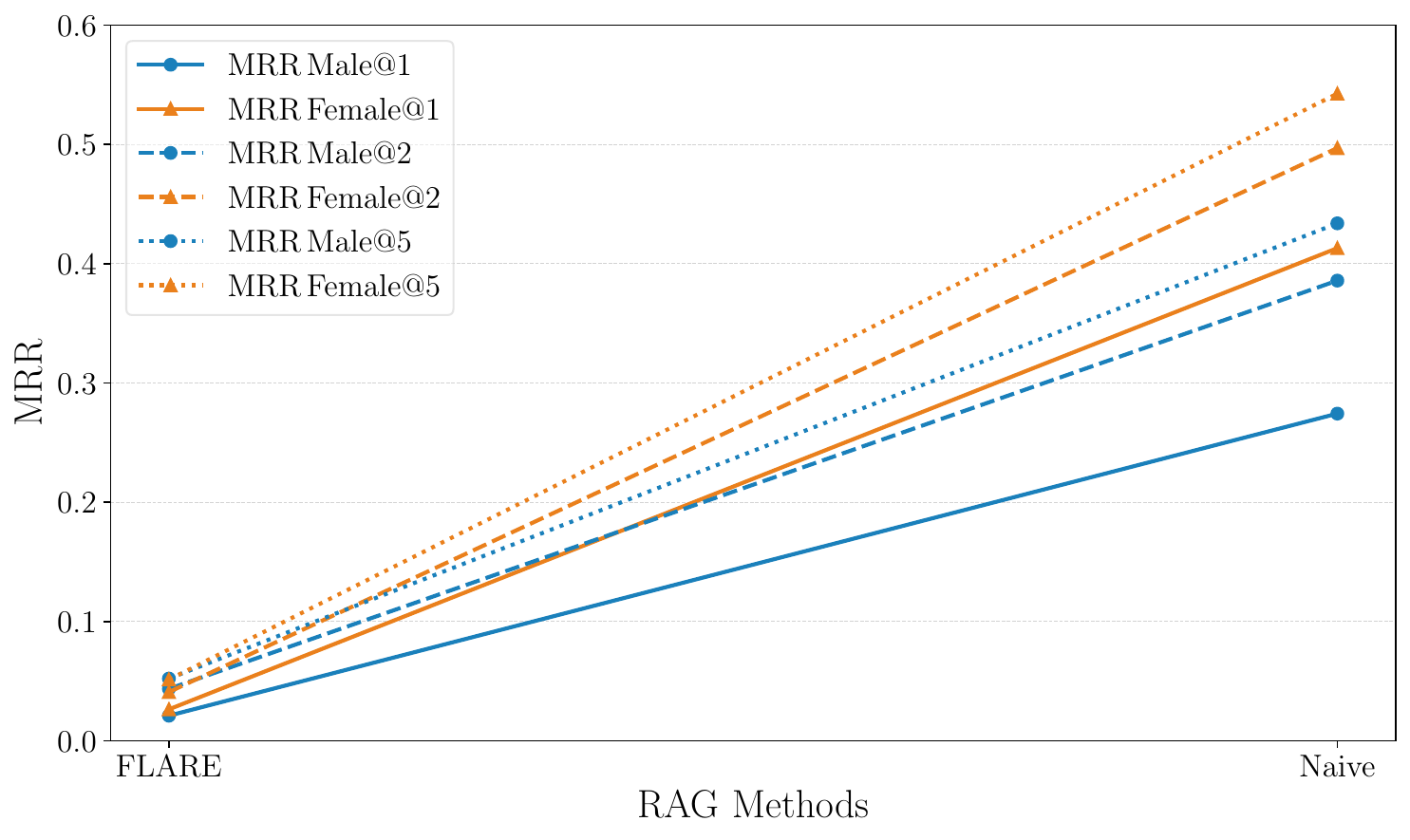}
  \caption {FLARE and Naive's MRR when retrieving 1, 2, and 5 documents using E5 in S1.}
  \label{fig:appendix-flare}
\end{figure}

Flare’s stability in EM and $\text{GD}_{\text{S1}}$ remains consistent regardless of the number of retrieved documents, showing performance similar to the Zero-Shot method (Figure \ref{fig:ret_num}). This is because Flare consistently retrieves very few golden documents, as reflected in its low MRR scores for both males and females (Figure \ref{fig:appendix-flare}). Consequently, its retrieval mechanism seems to have minimal impact on performance, which explains why its EM and $\text{GD}_{\text{S1}}$ remain stable even as more documents are retrieved. This stability likely stems from Flare’s retrieval approach, where it only retrieves documents when it detects uncertainty during generation, typically with low-confidence tokens. As a result, Flare retrieves fewer but highly specific documents, and its reliance on iteratively regenerating sentences without always requiring new documents further contributes to its stable performance. In contrast, the Naive method shows significant improvements in both EM and fairness (Figure \ref{fig:ret_num}) as it retrieves more documents. The Naive method's increasingly higher MRR scores for both males and females (Figure \ref{fig:appendix-flare}) indicates that the Naive method consistently retrieves more golden documents, which allows it to leverage the retrieval process more effectively, improving EM and decreasing unfairness.

% \subsection{retriever - bm25 vs e5}
% Key Insights on MRR and Unfairness:
% [
% Low MRR values (as seen with BM25) amplify the impact of any difference between female and male MRR on the unfairness ratio. Since BM25's MRR female@all is significantly larger than MRR male@all, this results in a strong female-favoring bias.
% Higher MRR values (as seen with E5), even when MRR female@all is still larger than MRR male@all, reduce the influence on unfairness. The larger MRR values for both genders provide a more balanced retrieval, minimizing the impact on fairness.
% EM and fairness tend to be more stable and balanced with E5, whereas BM25 introduces more variability in both EM and unfairness, leading to a stronger bias favoring females.
% ]

\end{document}